\PassOptionsToPackage{numbers,compress}{natbib}
\documentclass{article}

\usepackage[preprint]{neurips_2025}

\usepackage[utf8]{inputenc}
\usepackage[T1]{fontenc}
\usepackage{hyperref}
\usepackage{url}
\usepackage{booktabs}
\usepackage{amsfonts}
\usepackage{amsmath,amssymb,mathtools}
\usepackage{nicefrac}
\usepackage[table]{xcolor}

\definecolor{TabHeader}{HTML}{D9E4F2}   
\definecolor{TabGroup}{HTML}{EEF2F7}    
\definecolor{TabOffline}{HTML}{F4EEF9}  
\definecolor{TabSingle}{HTML}{EAF4FF}   
\definecolor{TabMulti}{HTML}{EAF8F1}    

\arrayrulecolor{black}

\usepackage{graphicx}
\usepackage{multirow}
\usepackage{float}
\usepackage{capt-of}
\usepackage{fvextra}
\usepackage{enumitem}
\usepackage[capitalize,noabbrev]{cleveref}

\DefineVerbatimEnvironment{PromptBox}{Verbatim}{
  fontsize=\footnotesize,
  breaklines=true,
  breakanywhere=true,
  frame=single,
  framesep=2mm
}

\setlist[itemize]{leftmargin=1.3em,itemsep=2pt,topsep=3pt,parsep=0pt,partopsep=0pt}
\setlist[enumerate]{leftmargin=1.5em,itemsep=2pt,topsep=3pt,parsep=0pt,partopsep=0pt}
\definecolor{TableHeader}{RGB}{244,247,252}
\definecolor{TableSection}{RGB}{232,239,251}
\definecolor{TableHighlight}{RGB}{240,245,255}

\title{Think While Watching: Online Streaming Segment-Level Memory for Multi-Turn Video Reasoning in Multimodal Large Language Models}

\author{
Lu Wang$^{1}$ \quad
Zhuoran Jin$^{1}$ \quad
Yupu Hao$^{1}$ \quad
Yubo Chen$^{1}$ \\
Kang Liu$^{1}$ \quad
Yulong Ao$^{2}$ \quad
Jun Zhao$^{1}$ \\
$^{1}$The Key Laboratory of Cognition and Decision Intelligence for Complex Systems,\\
Institute of Automation, Chinese Academy of Sciences, Beijing, China \\
$^{2}$Beijing Academy of Artificial Intelligence (BAAI), Beijing, China \\
\texttt{wanglu2026@ia.ac.cn, zhuoran.jin@nlpr.ia.ac.cn, haoyupu2023@ia.ac.cn} \\
\texttt{yubo.chen@nlpr.ia.ac.cn, kliu@nlpr.ia.ac.cn} \\
\texttt{aoyulong@outlook.com, jzhao@nlpr.ia.ac.cn}
}

\begin{document}

\maketitle

\begin{abstract}
Multimodal large language models (MLLMs) have demonstrated strong performance in offline video understanding tasks, but most remain constrained to offline inference or exhibit weak online reasoning ability, rendering online multi-turn interaction over continuously arriving video streams challenging. Existing streaming approaches adopt an interleaved perception-generation paradigm, which precludes concurrent perception and generation and induces early memory decay with growing streams, degrading long-range dependency modeling. We propose \textbf{Think While Watching}, a memory-anchored streaming video reasoning framework that maintains continuous segment-level memory during multi-turn interaction. We construct a three-stage, multi-round, chain-of-thought (CoT) dataset with a stage-matched training strategy while enforcing strict causality in streaming reasoning via a segment-level streaming causal mask and streaming positional encoding. At inference, we design an efficient pipeline that overlaps watching and thinking and adaptively selects the optimal attention backend. We evaluate our method under single-round and multi-round streaming input protocols. Based on Qwen3-VL, we improve single-round accuracy by 2.6\% on StreamingBench and 3.79\% on OVO-Bench. In the multi-round protocol, we maintain performance while reducing output tokens by 56\%. Our code is available at \href{https://github.com/wl666hhh/Think_While_Watching/}{GitHub}.
\end{abstract}

\section{Introduction}

\begin{figure*}[t]
  \centering
  \includegraphics[width=\linewidth]{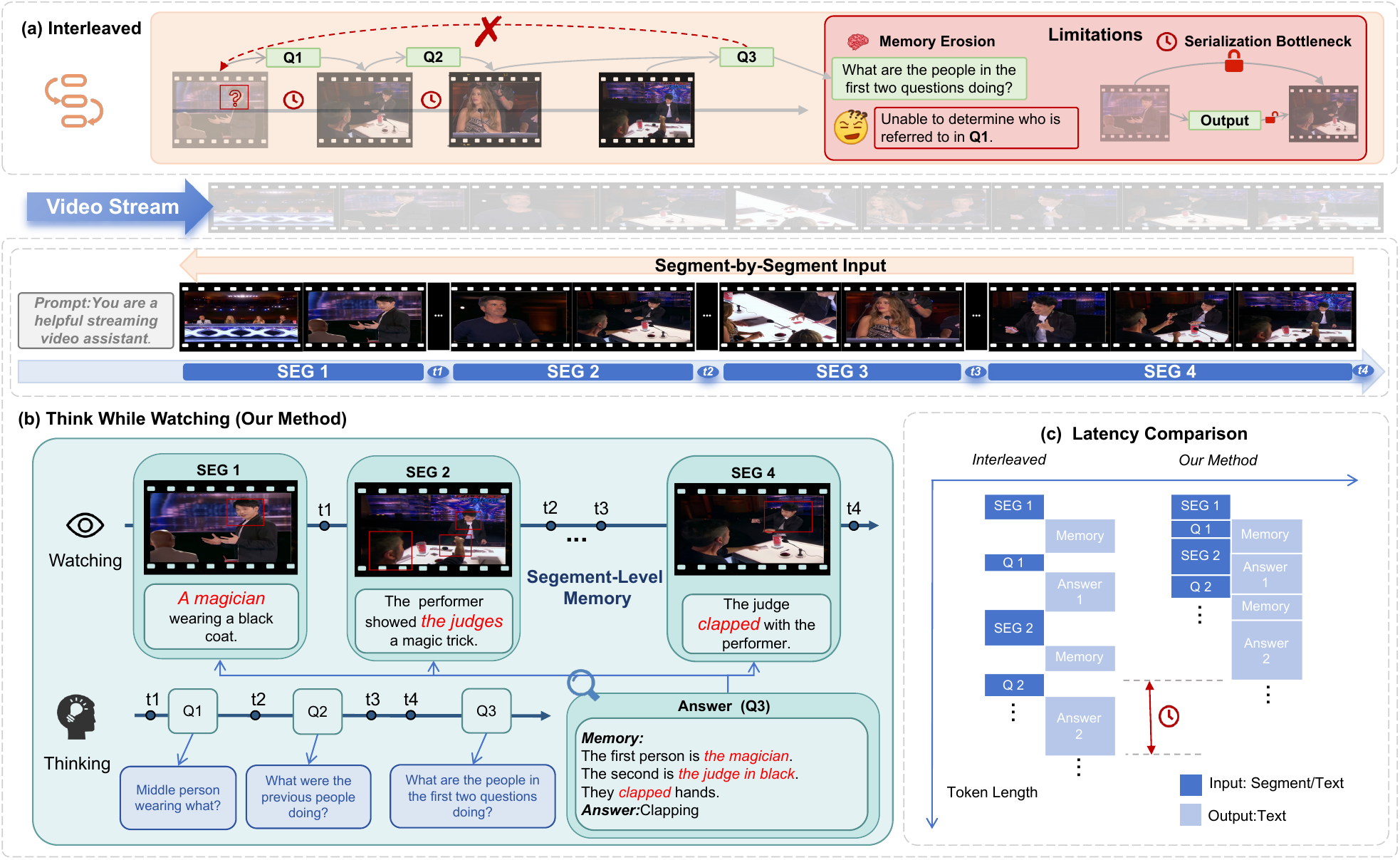}
  \caption{\textbf{Overview of Think While Watching.}
  \textbf{(a) Interleaved baseline.} Video perception and answer generation are executed sequentially, which can cause \textbf{memory erosion}, where early memory is forgotten, and a \textbf{serialization bottleneck}, where generation stalls further input ingestion.
  \textbf{(b) Think While Watching (ours).} The video frames are processed in segments (SEG 1 to SEG 4) to build a continuous \textbf{segment-level memory}. During streaming, questions are answered online by retrieving implicitly relevant memories while continuing to watch.
  \textbf{(c) Latency comparison.} A schematic timeline showing that interleaved processing accumulates queueing delay, while our decoupled design parallelizes segment processing and answering to reduce latency.}
  \label{fig:main}
\end{figure*}

Video understanding and reasoning are becoming central capabilities for multimodal assistants. While multimodal large language models (MLLMs) have achieved strong performance in offline video benchmarks where the full video is available before inference in a single-turn setting \cite{DBLP:journals/corr/abs-2511-04570,DBLP:journals/corr/abs-2508-04416,DBLP:journals/corr/abs-2503-21776,DBLP:journals/corr/abs-2501-03230}, many high-impact scenarios are inherently \textbf{streaming}: live broadcasting \cite{DBLP:journals/corr/abs-2511-05299,DBLP:conf/cvpr/ChenZLLMS25}, monitoring, robotic assistants \cite{wang2025streameqastreamingvideounderstanding}, and other streaming scenarios. In these settings, users may ask questions at any time, and the assistant must answer in real time while staying faithful to the visual evidence observed so far, especially under \textbf{multi-turn} interaction, where later questions often depend on earlier memories.

A widely adopted approach for streaming MLLMs is to interleave perception and generation \cite{DBLP:conf/cvpr/ChenLWLSGLGMS24,DBLP:journals/corr/abs-2412-08646,DBLP:conf/cvpr/ChenZLLMS25}. Although this reduces delay compared to offline approaches, it remains fundamentally serialized: text decoding blocks further video ingestion. This interleaved pattern leads to two phenomena.
First, \textbf{Memory Erosion}: multi-turn subsequent questions frequently refer back to earlier questions or earlier visual cues, but interleaving with generation tends to erode long-range capability. The failure case in Fig.~\ref{fig:main}(a) makes this explicit: a later query about the first two questions becomes unanswerable because the model forgets who Q1 refers to. This issue is also validated by our experimental results: for the Qwen3-VL-4B Thinking model, accuracy in the online multi-round setting drops $40.39\%$ compared to the offline setting, highlighting the severe challenge of maintaining long-term temporal consistency.
Second, \textbf{Serialization Bottleneck}: as illustrated on the top right of Fig.~\ref{fig:main}(a), once the model starts generation, the decoder effectively locks the streaming sequence, directly harming responsiveness in dynamic streams. The root cause is that autoregressive models use unified positional encoding, so new inputs must align with generated outputs whose length is unknown, forcing ingestion to pause and causing a serialization bottleneck. Fig.~\ref{fig:main}(c) further visualizes this effect: under interleaved processing, as the number of rounds accumulates, the input keeps piling up, leading to increasing end-to-end latency. To mitigate \textbf{Memory Erosion}, we make memory writing an explicit online behavior: for each observed segment, the model writes a memory note and appends it to a memory bank; when a question arrives, the model answers by implicitly integrating the relevant notes via the attention mechanism in Fig.~\ref{fig:main}(b). To break the \textbf{Serialization Bottleneck}, we assign independent positional encodings to decouple input and output streams at inference time, enabling input-output parallelism so the model can keep watching while thinking and thus reduce latency.

We propose \textbf{Think While Watching}, a memory-anchored streaming video reasoning framework for online multi-turn interaction. We represent a continuously arriving video as a sequence of segments and maintain a persistent \textbf{segment-level memory} throughout the dialogue. To make the framework practical, we design corresponding training and inference procedures. On the training side, we construct a three-stage, multi-round chain-of-thought (CoT) dataset with training matched to each stage, together with a \textbf{segment-level streaming causal mask} and \textbf{streaming positional encoding}, which jointly enforce strict causality throughout streaming reasoning. On the inference side, we design an efficient pipeline that overlaps watching with thinking. Our implementation is inspired by CPU process scheduling \cite{DBLP:books/wi/SGG2018} in operating systems: we organize inference as a \textbf{multi-stage pipeline} as illustrated in Fig.~\ref{fig:main}(c) and decouple continuous visual ingestion from text decoding via a dual KV cache \cite{DBLP:conf/acl/TongFLFZSS25,DBLP:journals/corr/abs-2510-17238,lin2026speakwatchingunleashingtrue}, enabling parallelism between perception and generation and mitigating serialization.

We evaluate Think While Watching under two streaming input protocols: \textbf{single-round}, where the input contains many arriving segments but the assistant answers one question, and \textbf{multi-round}, where the input contains many arriving segments and the assistant answers multiple questions over time. In the Qwen3-VL family, we improve single-round accuracy by 2.6\% on StreamingBench and 3.79\% on OVO-Bench. In the multi-round protocol, we preserve performance while reducing output tokens by 56\%.

\noindent\textbf{Our contributions} are as follows:
\begin{itemize}
  \item We propose \textbf{Think While Watching}, which maintains \textbf{segment-level memory} as a persistent state and answers each query by implicitly retrieving and integrating relevant memories, improving multi-turn consistency and enabling decoupled perception and generation. We further provide a practical \textbf{training and inference} pipeline with \textbf{three-stage training}, \textbf{streaming segment masking}, and \textbf{streaming positional encoding} for causal segment-level modeling, and a dual KV cache at inference time to support parallelism between perception and generation.
  \item We construct a three-stage, stage-aligned streaming CoT dataset with multi-round dialogues to support the proposed training strategy.
  \item On Qwen3-VL, we improve single-round accuracy by 2.6\% on StreamingBench and 3.79\% on OVO-Bench, while in multi-round streaming, we reduce output tokens by 56\% without accuracy drop.
\end{itemize}

\section{Related Work}

\subsection{Offline Video Understanding}
Offline video MLLMs are improved by structured perception and cognition pipelines and temporal reasoning designs \cite{DBLP:journals/corr/abs-2501-03230,DBLP:journals/corr/abs-2511-04570,DBLP:journals/corr/abs-2508-04416}, and by reinforcement learning for complex temporal reasoning \cite{DBLP:journals/corr/abs-2503-21776,DBLP:journals/corr/abs-2507-07966}. Most of these methods assume the full video is available before answering, leaving causal online multi-turn interaction less explored.

\subsection{Online Streaming Video Understanding}

\noindent\textbf{Benchmarks.}
StreamingBench evaluates the gap between offline models and streaming video understanding \cite{lin2024streamingbenchassessinggapmllms}, while OVO-Bench emphasizes real-world online video understanding \cite{DBLP:conf/cvpr/NiuLMGZHDDD0ZZC25}. Recent work further studies streaming along with active perception and multi-turn interaction \cite{Yang2025ViSPeak,lee2025streamgazegazeguidedtemporalreasoning,DBLP:journals/corr/abs-2510-14560,DBLP:conf/iclr/YangHDXQW0DX25,DBLP:journals/corr/abs-2505-02064}.

\noindent\textbf{Interleaved perception and generation.}
Many streaming systems alternate visual ingestion and text decoding, as in VideoLLM-online \cite{DBLP:conf/cvpr/ChenLWLSGLGMS24} and StreamChat \cite{DBLP:journals/corr/abs-2412-08646}, and scale streaming supervision for online interaction \cite{DBLP:conf/cvpr/ChenZLLMS25,DBLP:journals/corr/abs-2510-09608}. This coupling limits input-output parallelism and makes it harder to model dependencies over a long horizon across multiple turns.

\noindent\textbf{Memory and efficient inference for long-horizon streaming.}
For efficiency, one line reduces redundant visual tokens in streaming videos \cite{DBLP:journals/corr/abs-2504-17343,DBLP:conf/cvpr/ZhouABYMXNS24,DBLP:journals/corr/abs-2510-18269,DBLP:conf/iclr/XiongYYZ0ZL25}. Another line reuses historical context via KV cache retrieval and compression \cite{kim2025vrexrealtimestreamingvideo,DBLP:journals/corr/abs-2511-07278,DBLP:conf/iclr/DiYZLZCLHSJ25,DBLP:journals/corr/abs-2505-15269,DBLP:journals/corr/abs-2505-13140,DBLP:journals/corr/abs-2506-23825}. Persistent memory and long-term multimodal agent memory further support evidence reuse across long streams \cite{DBLP:journals/corr/abs-2509-24871,DBLP:journals/corr/abs-2508-09736,DBLP:journals/corr/abs-2508-01875}. Our work emphasizes stable segment-level memory for multi-turn streaming and an inference design that keeps perception and generation decoupled.

\section{Preliminary}
\label{sec:paradigm}

This section introduces the online multi-turn streaming video question answering setting studied in this work. A video is observed sequentially as a stream of segments, while a user may ask questions at arbitrary segment boundaries. The central requirement is strict streaming causality: at each turn, the system must produce its response using only the video content observed so far and the dialogue history, without accessing any future segments.

\subsection{Streams and Turns}
\label{sec:paradigm:stream}

\noindent\textbf{Segmented stream.}
We represent a video stream as an ordered sequence of segments
\begin{equation}
\mathbf{S}_{1:T} \triangleq \langle \mathbf{S}_1,\ldots,\mathbf{S}_T\rangle ,
\end{equation}
where each $\mathbf{S}_t$ denotes a contiguous chunk of frames. Segments arrive in temporal order, and the system processes them online.

\noindent\textbf{Multi-turn questioning.}
We consider an interaction with $R$ question and answer turns. At turn $r \in \{1,\ldots,R\}$, the user asks a question $q_r$ after the system has observed a prefix of the stream. Let $\tau_r \in \{1,\ldots,T\}$ denote the index of the latest observed segment when $q_r$ is issued. Equivalently, $q_r$ is asked after ingesting the segment prefix
\begin{equation}
\mathbf{S}_{1:\tau_r} \triangleq \langle \mathbf{S}_1,\ldots,\mathbf{S}_{\tau_r}\rangle .
\end{equation}
Since questions arrive over time, the indices are nondecreasing:
\begin{equation}
1 \le \tau_1 \le \tau_2 \le \cdots \le \tau_R \le T .
\end{equation}
The dialogue history before turn $r$ is
\begin{equation}
\mathcal{H}_{r-1} \triangleq \langle \langle q_1,a_1\rangle,\ldots,\langle q_{r-1},a_{r-1}\rangle\rangle .
\end{equation}
Under strict causality, the answer $a_r$ at turn $r$ is conditioned only on the observed video prefix $\mathbf{S}_{1:\tau_r}$, the question $q_r$, and the dialogue history $\mathcal{H}_{r-1}$.

\subsection{Streaming Protocols}
\label{sec:paradigm:protocols}

We consider two online evaluation protocols that share the same segmented stream $\mathbf{S}_{1:T}$ but differ in the number of question turns.

\noindent\textbf{Single-round streaming.}
Only one question is asked, so $R=1$. The system processes segments online and produces a single output for the question asked at $\tau_1$. We denote the model output as $\langle \pi_1, a_1\rangle$, where $\pi_1$ is an optional intermediate rationale such as chain of thought and $a_1$ is the final answer.

\noindent\textbf{Multi-round streaming.}
Multiple questions are asked, so $R>1$, at different times with nondecreasing $\tau_r$. At each turn $r$, the system must answer online using only the stream prefix $\mathbf{S}_{1:\tau_r}$ and the dialogue history $\mathcal{H}_{r-1}$, producing an output pair $\langle \pi_r, a_r\rangle$.

\subsection{Streaming Unit Notation}
\label{sec:paradigm:notation}

To describe training and inference in a single causal formulation, we serialize a streaming interaction as an interleaved sequence of received units and a one-to-one aligned sequence of generated units.

\noindent\textbf{Received units.}
Let the received unit sequence be
\begin{equation}
\mathbf{R}_{1:U} \triangleq \langle R_1,\ldots,R_U\rangle ,
\end{equation}
where each $R_u$ is either a visual segment unit $S_t$ that contains the content of $\mathbf{S}_t$, or a question unit $Q_r$ that contains the text $q_r$. We write $R_u \in \{S,Q\}$ to indicate the unit type. Let $\operatorname{idx}\lbrack \cdot \rbrack$ return the arrival index in $\mathbf{R}_{1:U}$, so $\operatorname{idx}\lbrack S_t \rbrack$ is the index $u$ where segment $t$ appears, and $\operatorname{idx}\lbrack Q_r \rbrack$ is the index $u$ where question $r$ appears.

\noindent\textbf{Generated units.}
For each received unit $R_u$, the model generates exactly one output unit $C_u$ in the same order, forming
\begin{equation}
\mathbf{C}_{1:U} \triangleq \langle C_1,\ldots,C_U\rangle .
\end{equation}
If $R_u=S_t$, then $C_u$ is a memory note denoted $m_t$. If $R_u=Q_r$, then $C_u$ is the question answering output that contains the rationale $\pi_r$ and answer $a_r$.

\noindent\textbf{Token lengths and visual grids.}
For any text unit $Y$ in $\{Q_1,\ldots,Q_R,C_1,\ldots,C_U\}$, let $L\lbrack Y \rbrack$ denote its text token length. For any segment unit $S_t$, let its visual token grid sizes be $\langle T_t,H_t,W_t\rangle$. Here $T_t$ is the number of visual tokens along the temporal axis, $H_t$ is the height axis, and $W_t$ is the width axis, defined by the vision encoder token grid for this segment. We will also use a unit span function $\Delta\lbrack R_u \rbrack$ that assigns each received unit a nonoverlapping input position span:
\begin{equation}
\Delta\lbrack R_u \rbrack =
\begin{cases}
\max\{T_u,H_u,W_u\}, & R_u \in \{S\},\\
L\lbrack R_u \rbrack, & R_u \in \{Q\}.
\end{cases}
\label{eq:delta}
\end{equation}

\section{Method}

\begin{figure*}[t]
  \centering
  \includegraphics[width=\linewidth]{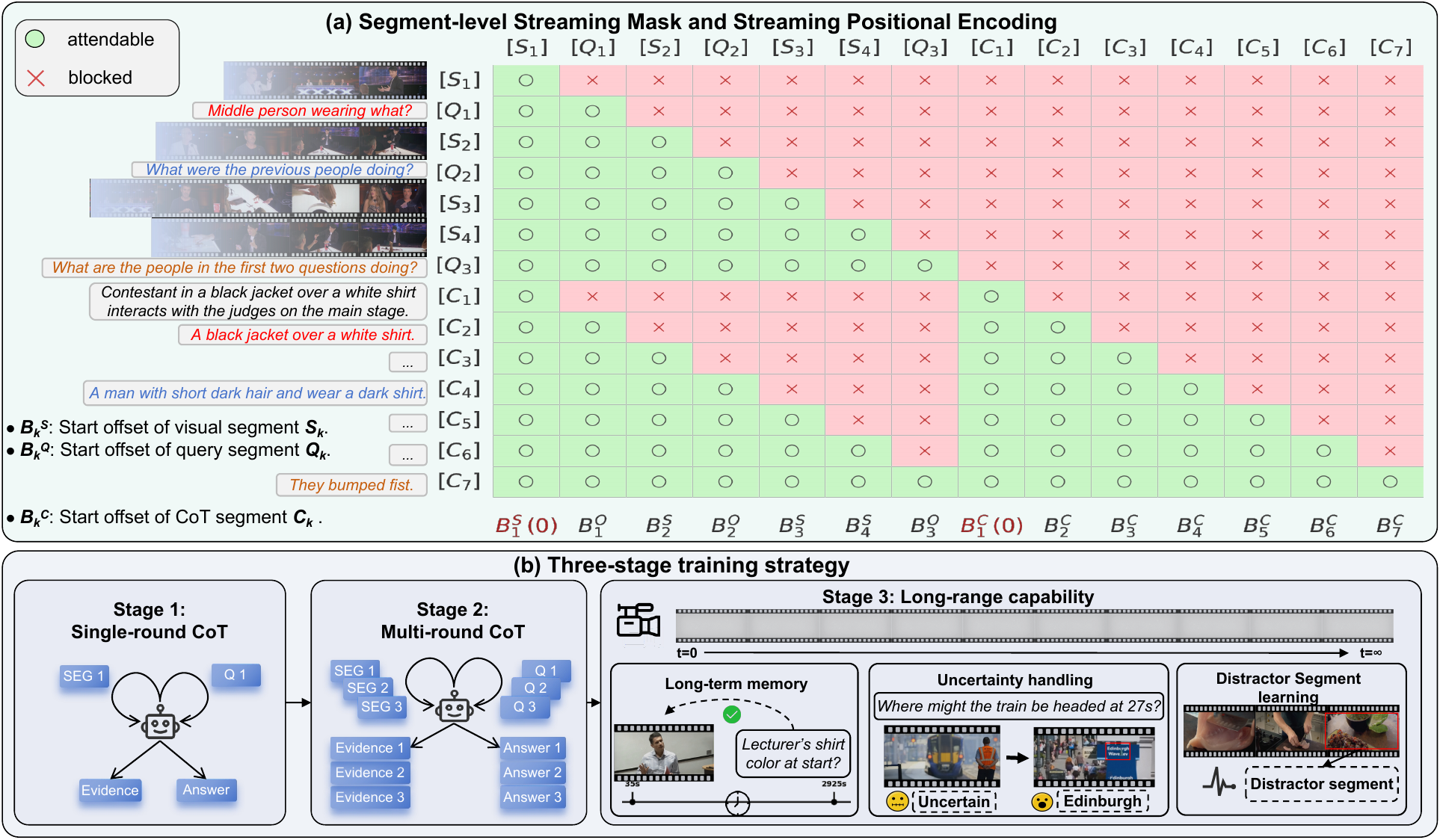}
  \caption{\textbf{Training components of Think While Watching.}
  (a) segment-level streaming attention mask and streaming positional encoding: example input stream $\mathbf{R}=\langle S_1,Q_1,S_2,Q_2,S_3,S_4,Q_3\rangle$ with generated outputs $\mathbf{C}=\langle C_1,\ldots,C_7\rangle$. Green indicates the source prefix available at time step $u$, which $C_u$ is allowed to attend to. Red masks all future segments to prevent information leakage. For positional encoding, we use separate position indices for the input and output streams.
  (b) Three-stage training strategy: single-round CoT for streaming input adaptation, multi-round CoT for multi-turn interaction, and long-range capability training for long-term memory, uncertainty handling, and distractor learning.}
  \label{fig:train}
\end{figure*}

A simple streaming design interleaves perception and generation but is inherently serial: autoregressive decoding halts further input ingestion, and the interleaving pattern mismatches the LLM pretraining format. Think While Watching generates segment-level memory notes online and decouples perception and generation, enabling interaction in real time across multiple turns. Details are shown in Fig.~\ref{fig:train}.

\subsection{Segment-level Memory Notes}
\label{sec:method:memory}

To mitigate Memory Erosion, our method maintains a persistent segment-level memory bank as the online state for multi-turn streaming. For each arriving segment $\mathbf{S}_t$, the model writes exactly one memory note and appends it to the bank. The memory bank after observing the segment prefix $\mathbf{S}_{1:t}$ is defined as
\begin{equation}
\mathcal{M}_t \triangleq \{\langle i, m_i\rangle\}_{i=1}^{t}.
\end{equation}
Each note $m_t$ is a compact text unit grounded in $\mathbf{S}_t$ that records reusable evidence, including key entities and attributes, salient actions and interactions, scene changes, and short-range temporal relations. We denote the memory-writing function implemented by the MLLM backbone with parameters $\theta$ as $\mathrm{Mem}_\theta[\cdot]$, and write
\begin{equation}
m_t = \mathrm{Mem}_\theta[\mathbf{S}_t],
\qquad
C_{\operatorname{idx}[S_t]} = m_t .
\end{equation}

\noindent When a question $q_r$ is issued after observing segment $\tau_r$, the model answers by conditioning on the current question, the dialogue history, and the available memory prefix, while letting attention implicitly select relevant notes:
\begin{equation}
\langle \pi_r, a_r\rangle \sim
p_\theta\!\left[\pi_r, a_r \mid q_r, \mathcal{H}_{r-1}, \mathcal{M}_{\tau_r}\right].
\end{equation}

\subsection{Streaming Architecture}
\label{sec:method:train}

\noindent\textbf{Prefix and suffix formatting.}
To match the standard SFT format of LLMs, we serialize each training example as a source prefix followed by a target suffix. The source prefix contains the entire received unit stream $\mathbf{R}_{1:U}$, while the target suffix contains the aligned generated stream $\mathbf{C}_{1:U}$. Without additional constraints, this serialization would leak future received units to earlier generated units.

\noindent\textbf{Segment-level streaming attention mask.}
We feed the concatenated sequence
$\langle R_1,\ldots,R_U,\, C_1,\ldots,C_U\rangle$ with a segment-level mask $M^{\mathrm{seg}}$ to enforce streaming causality. Let $A$ denote the segment that contributes query tokens and $B$ denote the segment that contributes key and value tokens, with
$A,B \in \{R_1,\ldots,R_U,C_1,\ldots,C_U\}$.
The mask is defined as:
\begin{equation}
M^{\mathrm{seg}}\lbrack A,B\rbrack=
\begin{cases}
\mathbb{I}\lbrack v \le u\rbrack, & A=R_u,\; B=R_v,\\
\mathbb{I}\lbrack v \le u\rbrack, & A=C_u,\; B=R_v,\\
\mathbb{I}\lbrack k \le u\rbrack, & A=C_u,\; B=C_k,\\
0, & \text{otherwise.}
\end{cases}
\label{eq:seg-mask}
\end{equation}
Here $u$ is the arrival index of the querying unit $A$. For the attended unit $B$, we use $v$ if $B$ is a received unit $R_v$ and $k$ if $B$ is a generated unit $C_k$. The first three cases in Eq.~\eqref{eq:seg-mask} enforce streaming causality: the received stream is causal in arrival order, each generated unit $C_u$ can attend to the received prefix up to step $u$, and generated units are causal with access only to $C_{1:u}$. All remaining connections are masked, including $R_u \rightarrow C_k$ and $C_u \rightarrow R_v$ for $v>u$. We obtain token-level masks by expanding $M^{\mathrm{seg}}$ and applying standard causal masking within each $C_u$. As shown in Fig.~\ref{fig:train}(a), $C_1$ attends only to $S_1$, $C_2$ attends to $\langle S_1,Q_1\rangle$, and $C_3$ attends to $\langle S_1,Q_1,S_2\rangle$.

\noindent\textbf{Streaming positional encoding with MRoPE.}
We build on Multimodal Rotary Positional Embeddings MRoPE~\cite{DBLP:journals/corr/abs-2409-12191}, but decouple the input and output to support parallel reasoning. Specifically, the input stream follows the standard cumulative offset scheme, while the output stream independently starts positional encoding from $0$. We use $B$ to represent the base offset and compute the start offsets of the $k$-th visual segment $S_k$ input, the $k$-th question $Q_k$ input, and the $k$-th generated unit $C_k$ output:
\begin{equation}
\label{eq:stream-mrope}
B_k =
\left\{
\begin{aligned}
B_k^{S} &= \sum_{u < \operatorname{idx}\!\left[ S_k \right]} \Delta\!\left[ R_u \right],\\
B_k^{Q} &= \sum_{u < \operatorname{idx}\!\left[ Q_k \right]} \Delta\!\left[ R_u \right],\\
B_k^{C} &=
\begin{cases}
0, & k=1,\\
\sum_{i=1}^{k-1} L\!\left[ C_i \right], & k\ge 2.
\end{cases}
\end{aligned}
\right.
\end{equation}

\noindent In this design, $B_k^{S}$ and $B_k^{Q}$ are computed only from the received input prefix, while $B_k^{C}$ is computed only from previously generated tokens. Therefore, even when the output length is still unknown during decoding, newly arriving input segments can always be assigned correct input positions.\footnote{MRoPE extends RoPE to multimodal tokens by applying rotary positional encoding along modality-specific axes. For a visual segment unit $S_k$ with token grid size $\langle T_k,H_k,W_k\rangle$, a token at local grid coordinate $\langle t,h,w\rangle$ is assigned global coordinates $\langle t+B_k^{S},\,h+B_k^{S},\,w+B_k^{S}\rangle$, where the start offset $B_k^{S}$ is given in Eq.~\eqref{eq:stream-mrope}. For a text unit $Y\in\{Q_k,C_k\}$ with length $L[Y]$, a token at local position $n$ uses $n+B_k^{Q}$ for question inputs and $n+B_k^{C}$ for generated outputs. The contribution of each received unit to the input position budget is determined by the unit span $\Delta[R_u]$ in Eq.~\eqref{eq:delta}.}

\subsection{Streaming Training}
\label{sec:method:data}

\noindent\textbf{Three-stage training.}
We fine-tune the MLLM in three stages: Stage~1 learns to write segment-level memory notes and answer single-round queries. Stage~2 scales to multi-round dialogues. Stage~3 targets long-range behaviors on long videos, including long-term evidence recall, uncertainty handling, and distractor segment learning where we insert irrelevant frames as distractors. In Fig.~\ref{fig:train}(b), Stage~3 covers three long-horizon behaviors: \textbf{long-term memory} for recalling early evidence in late queries, \textbf{uncertainty handling} for deferring commitment when evidence is not yet observable, and \textbf{distractor robustness} for ignoring irrelevant segments during streaming.

\noindent\textbf{Three-stage Streaming CoT Dataset Generation.}
Streaming CoT \cite{DBLP:journals/corr/abs-2510-25332} datasets for MLLMs are extremely scarce. Multi-round streaming CoT with memory notes is largely absent. We therefore synthesize a three-stage dataset that matches our three-stage training.

\noindent\textbf{Stage 1 and Stage 2 short video streaming CoT.}
We use VideoChatOnline-IT \cite{huang2024online} as the source pool and sample up to 64 frames per instance. Stage~1 constructs 5,160 single-round instances from temporal perception subsets. Stage~2 converts 8,513 short video QA instances into 2,752 multi-round dialogues by grouping questions over the same video prefix. For both stages, we use GPT-5.2 to generate memory-anchored CoT based on the original dataset QAs.

\noindent\textbf{Stage 3 long-range streaming CoT.}
We collect long videos from YouTube using 500 keywords spanning three categories: tutorial for procedural content, lecture for explanatory content, and longform for continuous recordings. We then generate 1,500 long video instances with balanced input lengths of 100 to 200 frames, 200 to 300 frames, and 300 or more frames, and each instance contains 3 to 5 rounds. QA and CoT generation follow the same procedure as in Stage~1 and Stage~2. Details of the dataset and the prompt can be found in Appendix~\ref{app:data}.

\noindent\textbf{Quality inspection.}
We enforce the core constraints in Table~\ref{tab:prompt-constraints} during synthesis, and additionally verify that each example contains exactly $S+Q$ output items.

\begin{table}[t]
\caption{\textbf{Dataset statistics across three training stages.} Stages 1\&2 are built from VideoChatOnline-IT short videos, while Stage~3 contains long-range multi-round dialogues from YouTube. Video duration is reported in seconds (min/avg/max).}
\label{tab:data}
\centering
\scriptsize
\setlength{\tabcolsep}{3.5pt}
\renewcommand{\arraystretch}{1.05}
\resizebox{\linewidth}{!}{
\begin{tabular}{@{}lcccccccc@{}}
\toprule
\textbf{Stage} & \textbf{Source} & \textbf{Instances} & \textbf{Rounds} & \textbf{Avg. rounds} & \textbf{Frames} & \textbf{Min (s)} & \textbf{Avg (s)} & \textbf{Max (s)} \\
\midrule
1 & VideoChatOnline-IT & 5,160 & 5,160 & 1.00 & $\le 64$ & 8.18 & 79.40 & 3550.10 \\
2 & VideoChatOnline-IT & 2,752 & 8,513 & 3.09 & $\le 64$ & 2.00 & 400.92 & 3443.97 \\
3 & YouTube & 1,500 & 6,000 & 4.00 & 100-300+ & 600.12 & 1697.30 & 3595.03 \\
\bottomrule
\end{tabular}}
\end{table}

\subsection{Streaming Inference}
\label{sec:method:infer}

\noindent\textbf{Parallel reasoning.}
For real-time deployment, we adopt a dual KV cache implementation that decouples continuous source ingestion from autoregressive decoding. This engineering pattern is common in recent streaming systems \cite{DBLP:journals/corr/abs-2510-17238}. Meanwhile, we keep the same segment-level streaming mask and streaming MRoPE at inference time to ensure consistency with training.

\noindent\textbf{Adaptive attention backend.}
During decoding, our streaming mask is not always a standard causal mask: queries must attend to an allowed source prefix while remaining causal over the generated suffix, so the query and key lengths can differ ($q_{\text{len}} \neq k_{\text{len}}$). We therefore choose the attention backend adaptively---using Flash Attention \cite{dao2022flashattentionfastmemoryefficientexact} when the pattern is standard causal, and otherwise applying an explicit streaming mask with memory-efficient attention \cite{rabe2022selfattentiondoesneedon2}. Specifically, we use Flash Attention for source prefilling ($q_{\text{len}} = k_{\text{len}}$) and for autoregressive steps ($q_{\text{len}} = 1$), and switch to memory-efficient attention when $1 < q_{\text{len}} < k_{\text{len}}$ to enforce the custom streaming mask. This preserves segment-level causality while keeping inference fast.

\section{Experiments}
\label{sec:experiments}

\subsection{Datasets and Setup}
\label{sec:exp_setup}

We evaluate online streaming performance on StreamingBench~\cite{lin2024streamingbenchassessinggapmllms} and OVO-Bench~\cite{DBLP:conf/cvpr/NiuLMGZHDDD0ZZC25}. StreamingBench focuses on streaming video understanding and includes four subsets. OVO-Bench emphasizes real-world video understanding under three subsets. More details of datasets are in Appendix~\ref{app:data}.

\noindent\textbf{Evaluation protocols.}
We evaluate models under both offline and online protocols. In the offline protocol, we adopt a Batch setting where all sampled frames from the entire video are provided as a single input, and the model answers the question after observing the complete video. In the online protocol, we consider single-turn and multi-turn interaction. For single-turn online evaluation, we segment each video according to the question timestamps provided by the benchmark, forming consecutive temporal segments $[0,t_1]$, $[t_1,t_2]$, and so on. If any segment lasts longer than $60$s, we further split it into $30$s chunks. For multi-turn evaluation, we use the same segmentation strategy, but the model must answer multiple questions online as segments continuously arrive.

\noindent\textbf{Backbones and checkpoints.}
We evaluate our method with Qwen3-VL backbones at 2B, 4B, and 8B scales. We use the Instruct model for training and compare its performance with the Thinking model. TWW is used to denote our method in the following. Stage~2 and Stage~3 refer to the checkpoints obtained after training up to the second and third stages.

\subsection{Baselines}
\label{sec:exp_baselines}

\noindent\textbf{Offline evaluation.}
We evaluate Gemini 1.5 Pro \cite{geminiteam2024gemini15unlockingmultimodal} and GPT-4o \cite{openai2024gpt4ocard} as representative closed-source MLLMs. Qwen3-VL-Instruct and Qwen3-VL-Thinking are used as open-source baselines. We also report results for our TWW$_{\text{Batch,S2}}$ and TWW$_{\text{Batch,S3}}$ checkpoints, corresponding to Stage~2 and Stage~3, evaluated under the offline batch protocol.

\noindent\textbf{Online evaluation.}
For online evaluation, Instruct$_{\text{online}}$ and Thinking$_{\text{online}}$ run Qwen3-VL-Instruct and Qwen3-VL-Thinking, respectively, under the multi-turn protocol in Sec.~\ref{sec:exp_setup}. We further evaluate our checkpoints under streaming settings: TWW$_{\text{single-turn,S2}}$ and TWW$_{\text{single-turn,S3}}$ follow the single-round streaming protocol, while TWW$_{\text{multi-turn,S2}}$ and TWW$_{\text{multi-turn,S3}}$ follow the multi-round streaming protocol. Finally, Interleaved alternates between ingesting one segment and decoding text, coupling perception and generation as a naive streaming baseline. More online baselines and details are in Appendix~\ref{app:baselines}.

\subsection{Metrics}
\label{sec:exp_metrics}

We use accuracy to evaluate performance on each benchmark and each evaluation regime. We also report Avg Tokens, the average number of generated output tokens per query. Token Reduce, denoted by $\Delta$\%, is the percentage reduction of Avg Tokens compared with the Thinking baseline of the same backbone size, and Avg Frames, the average number of processed frames per query. For latency, we report TTFT, time to first token, measured as the number of tokens processed before the first answer token is produced.

\subsection{Main Results}
\label{sec:exp_main}

\Cref{tab:streamingbench_main,tab:ovobench_main} report results on two streaming benchmarks, StreamingBench and OVO-Bench. We summarize the key findings below.

\begin{enumerate}[label=\textbf{\arabic*.}]
  \item \textbf{Naive streaming inference collapses without streaming-aligned training, highlighting the difficulty of multi-turn streaming.} On StreamingBench, directly running Instruct$_{\text{online}}$ and Thinking$_{\text{online}}$ achieves only 21.47\% and 18.13\% overall, compared with 56.67\% and 58.52\% with Qwen3-VL-4B in the offline batch setting. A similar drop is observed on OVO-Bench: 21.45\% and 16.21\% versus 50.32\% and 50.70\%, showing that multi-turn streaming is nontrivial and requires streaming-aligned supervision.

  \item \textbf{Streaming-aligned supervision improves accuracy.} With Qwen3-VL-4B, TWW$_{\text{single-turn,S3}}$ improves overall accuracy from 58.52\% to 60.04\% on StreamingBench and from 50.70\% to 55.02\% on OVO-Bench compared with the Thinking baseline.

  \item \textbf{Long-video training strengthens streaming behavior.} Stage~3 generally improves upon Stage~2. For example, on OVO-Bench with the 4B backbone, TWW$_{\text{single-turn,S3}}$ improves from 54.51\% to 55.02\%.

  \item \textbf{Multi-turn segment-level memory yields a strong accuracy--efficiency tradeoff.} Under the multi-turn protocol, TWW$_{\text{multi-turn,S3}}$ maintains competitive accuracy while substantially reducing decoding tokens. With the 4B backbone, it achieves 57.40\% on StreamingBench with an average of 302.56 tokens, reducing token usage by 56.10\%. On OVO-Bench, it obtains 51.80\% with 255.91 tokens on average, reducing token usage by 45.80\%.
\end{enumerate}
We further analyzed the types of errors in Appendix~\ref{app:error}.

\definecolor{TabHeader}{HTML}{D9E4F2}   
\definecolor{TabGroup}{HTML}{EEF2F7}    
\definecolor{TabOffline}{HTML}{F4EEF9}  
\definecolor{TabSingle}{HTML}{EAF4FF}   
\definecolor{TabMulti}{HTML}{EAF8F1}    

\newcommand{\offc}[1]{\cellcolor{TabOffline}{#1}}
\newcommand{\sinc}[1]{\cellcolor{TabSingle}{#1}}
\newcommand{\mulc}[1]{\cellcolor{TabMulti}{#1}}

\begin{table*}[t]
\caption{\textbf{StreamingBench results} with accuracy. Columns left of the double bar report \textbf{performance}, higher is better, while columns on the right report \textbf{efficiency} for open-source models only. $\Delta$ is computed against the same backbone Thinking baseline. Avg Frames is 148.35 for the single-turn protocol and 62.58 for the multi-turn protocol.}
\label{tab:streamingbench_main}
\centering
\scriptsize
\setlength{\tabcolsep}{3.0pt}
\renewcommand{\arraystretch}{1.11}
\resizebox{\textwidth}{!}{
\begin{tabular}{@{}llcccccc||cc@{}}
\toprule
& & \multicolumn{6}{c||}{\textbf{Performance}} & \multicolumn{2}{c}{\textbf{Efficiency}} \\
\cmidrule(lr){3-8}\cmidrule(lr){9-10}
\rowcolor{TabHeader}
\textbf{Regime} & \textbf{Method} &
\textbf{SQA}$\uparrow$ & \textbf{OmniSource}$\uparrow$ & \textbf{Realtime}$\uparrow$ & \textbf{Proactive}$\uparrow$ &
\textbf{Overall}$\uparrow$ & \textbf{$\Delta$}$\uparrow$ &
\textbf{Avg Tokens}$\downarrow$ & \textbf{Token Reduce}$\uparrow$ \\
\midrule

\rowcolor{TabGroup}
\multicolumn{10}{c}{\textbf{Closed-source models}} \\
\midrule
\multirow{2}{*}{Offline}
& Gemini 1.5 Pro\cite{geminiteam2024gemini15unlockingmultimodal} & 54.80 & 67.80 & 77.39 & 45.10 & 70.26 & - & - & - \\
& GPT-4o\cite{openai2024gpt4ocard} & 32.80 & 50.95 & 74.54 & 56.86 & 64.10 & - & - & - \\
\midrule

\rowcolor{TabGroup}
\multicolumn{10}{c}{\textbf{Open-source models}} \\
\midrule
\multirow{4}{*}{Online}
& Flash-VStream-7B\cite{DBLP:journals/corr/abs-2506-23825} & 26.80 & 26.00 & 23.23 & 1.96 & 24.04 & - & - & - \\
& VideoLLM-online-8B\cite{DBLP:conf/cvpr/ChenLWLSGLGMS24} & 30.80 & 28.45 & 35.99 & 3.92 & 32.48 & - & - & - \\
& Dispider-7B\cite{DBLP:conf/cvpr/0001DD0ZCL025} & 34.80 & 35.66 & 67.63 & 25.34 & 53.12 & - & - & - \\
& StreamAgent-7B\cite{DBLP:journals/corr/abs-2508-01875} & 39.60 & 36.26 & 74.28 & 28.90 & 57.02 & - & - & - \\
\midrule

\rowcolor{TabGroup}
\multicolumn{10}{c}{\textbf{Qwen3-VL-2B}} \\
\midrule
\multirow{4}{*}{Offline}
& Instruct & 37.60 & 31.60 & 68.36 & 29.60 & 52.24 & -1.17 & - & - \\
& Thinking & 34.00 & 31.73 & 70.02 & \textbf{36.80} & 53.41 & +0.00 & 1232.91 & 0.00 \\
& TWW$_{\text{Batch,S2}}$ & 44.00 & 33.13 & 69.16 & 33.20 & 53.75 & +0.34 & 1012.35 & 17.89 \\
& \offc{\textbf{TWW$_{\text{Batch,S3}}$}}
& \offc{\textbf{44.80}}
& \offc{\textbf{33.20}}
& \offc{\textbf{69.20}}
& \offc{36.00}
& \offc{\textbf{54.00}}
& \offc{\textbf{+0.59}}
& \offc{1102.34}
& \offc{10.59} \\
\midrule
\multirow{6}{*}{Online}
& Instruct$_{\text{online}}$ & 9.20 & 28.80 & 21.28 & 13.20 & 22.67 & -30.74 & - & - \\
& Thinking$_{\text{online}}$ & 8.40 & 11.60 & 12.92 & 19.20 & 12.58 & -40.83 & 832.23 & 32.50 \\
\cmidrule(lr){2-10}
& TWW$_{\text{single-turn,S2}}$ & 47.20 & 34.20 & 71.84 & 32.80 & 55.76 & +2.35 & 923.58 & 25.09 \\
& \sinc{\textbf{TWW$_{\text{single-turn,S3}}$}}
& \sinc{\textbf{48.00}}
& \sinc{\textbf{34.27}}
& \sinc{\textbf{72.00}}
& \sinc{34.40}
& \sinc{\textbf{56.00}}
& \sinc{\textbf{+2.59}}
& \sinc{930.23}
& \sinc{24.55} \\
\cmidrule(lr){2-10}
& TWW$_{\text{multi-turn,S2}}$ & 42.40 & 31.33 & 69.20 & 33.60 & 53.11 & -0.30 & \textbf{285.42} & \textbf{76.85} \\
& \mulc{\textbf{TWW$_{\text{multi-turn,S3}}$}}
& \mulc{45.20}
& \mulc{31.47}
& \mulc{69.24}
& \mulc{34.80}
& \mulc{\textbf{53.40}}
& \mulc{-0.01}
& \mulc{300.20}
& \mulc{75.65} \\
\midrule

\rowcolor{TabGroup}
\multicolumn{10}{c}{\textbf{Qwen3-VL-4B}} \\
\midrule
\multirow{4}{*}{Offline}
& Instruct & 37.20 & 38.47 & 71.36 & 38.40 & 56.67 & -1.85 & - & - \\
& Thinking & 46.40 & 36.53 & \textbf{74.50} & 42.80 & 58.52 & +0.00 & 689.22 & 0.00 \\
& TWW$_{\text{Batch,S2}}$ & 42.40 & 39.60 & 71.84 & 39.60 & 57.67 & -0.85 & 594.28 & 13.77 \\
& \offc{\textbf{TWW$_{\text{Batch,S3}}$}}
& \offc{44.00}
& \offc{\textbf{39.67}}
& \offc{71.88}
& \offc{40.80}
& \offc{57.87}
& \offc{-0.65}
& \offc{620.35}
& \offc{9.99} \\
\midrule
\multirow{6}{*}{Online}
& Instruct$_{\text{online}}$ & 12.80 & 21.53 & 22.88 & 15.60 & 21.47 & -37.05 & - & - \\
& Thinking$_{\text{online}}$ & 21.20 & 21.47 & 16.92 & 7.20 & 18.13 & -40.39 & 482.24 & 30.03 \\
\cmidrule(lr){2-10}
& TWW$_{\text{single-turn,S2}}$ & 46.00 & 40.67 & 74.36 & 41.20 & 59.71 & +1.19 & 558.12 & 19.02 \\
& \sinc{\textbf{TWW$_{\text{single-turn,S3}}$}}
& \sinc{\textbf{46.80}}
& \sinc{\textbf{41.00}}
& \sinc{74.48}
& \sinc{\textbf{43.20}}
& \sinc{\textbf{60.04}}
& \sinc{\textbf{+1.52}}
& \sinc{570.68}
& \sinc{17.20} \\
\cmidrule(lr){2-10}
& TWW$_{\text{multi-turn,S2}}$ & 40.80 & 39.20 & 71.20 & 40.00 & 57.11 & -1.41 & \textbf{291.86} & \textbf{57.65} \\
& \mulc{\textbf{TWW$_{\text{multi-turn,S3}}$}}
& \mulc{43.60}
& \mulc{39.33}
& \mulc{71.28}
& \mulc{40.80}
& \mulc{\textbf{57.40}}
& \mulc{-1.12}
& \mulc{302.56}
& \mulc{56.10} \\
\midrule

\rowcolor{TabGroup}
\multicolumn{10}{c}{\textbf{Qwen3-VL-8B}} \\
\midrule
\multirow{4}{*}{Offline}
& Instruct & 44.40 & 37.53 & 73.60 & 36.00 & 57.87 & -0.35 & - & - \\
& Thinking & 45.60 & 35.47 & 74.46 & \textbf{44.80} & 58.21 & +0.00 & 759.30 & 0.00 \\
& TWW$_{\text{Batch,S2}}$ & 52.80 & 38.93 & 74.56 & 38.00 & 59.44 & +1.23 & 708.16 & 6.70 \\
& \offc{\textbf{TWW$_{\text{Batch,S3}}$}}
& \offc{\textbf{53.20}}
& \offc{\textbf{39.07}}
& \offc{\textbf{74.64}}
& \offc{40.00}
& \offc{\textbf{59.67}}
& \offc{\textbf{+1.45}}
& \offc{720.30}
& \offc{5.10} \\
\midrule
\multirow{6}{*}{Online}
& Instruct$_{\text{online}}$ & 17.20 & 21.47 & 25.12 & 17.60 & 23.04 & -35.17 & - & - \\
& Thinking$_{\text{online}}$ & 14.80 & 13.00 & 17.56 & 16.40 & 15.82 & -42.39 & 573.75 & 24.41 \\
\cmidrule(lr){2-10}
& TWW$_{\text{single-turn,S2}}$ & 54.00 & 40.07 & 77.60 & 39.60 & 61.67 & +3.46 & 651.74 & 14.13 \\
& \sinc{\textbf{TWW$_{\text{single-turn,S3}}$}}
& \sinc{\textbf{54.40}}
& \sinc{\textbf{40.67}}
& \sinc{\textbf{77.68}}
& \sinc{41.60}
& \sinc{\textbf{62.04}}
& \sinc{\textbf{+3.83}}
& \sinc{660.92}
& \sinc{12.92} \\
\cmidrule(lr){2-10}
& TWW$_{\text{multi-turn,S2}}$ & 48.80 & 37.40 & 74.32 & 38.40 & 58.60 & +0.39 & \textbf{288.64} & \textbf{61.97} \\
& \mulc{\textbf{TWW$_{\text{multi-turn,S3}}$}}
& \mulc{50.00}
& \mulc{37.47}
& \mulc{74.40}
& \mulc{40.00}
& \mulc{\textbf{58.82}}
& \mulc{+0.61}
& \mulc{290.82}
& \mulc{61.68} \\
\bottomrule
\end{tabular}}
\end{table*}

\begin{table*}[!h]
\caption{\textbf{OVO-Bench results}. Avg Frames is 63.23 for the single-turn protocol and 25.47 for the multi-turn protocol.}
\label{tab:ovobench_main}
\centering
\scriptsize
\setlength{\tabcolsep}{2.8pt}
\renewcommand{\arraystretch}{1.11}
\resizebox{\textwidth}{!}{
\begin{tabular}{@{}llccccc||cc@{}}
\toprule
& & \multicolumn{5}{c||}{\textbf{Performance}} & \multicolumn{2}{c}{\textbf{Efficiency}} \\
\cmidrule(lr){3-7}\cmidrule(lr){8-9}
\rowcolor{TabHeader}
\textbf{Regime} & \textbf{Method} &
\textbf{Backward}$\uparrow$ & \textbf{Realtime}$\uparrow$ & \textbf{Forward}$\uparrow$ & \textbf{Overall}$\uparrow$ & \textbf{$\Delta$}$\uparrow$ &
\textbf{Avg Tokens}$\downarrow$ & \textbf{Token Reduce}$\uparrow$ \\
\midrule

\rowcolor{TabGroup}
\multicolumn{9}{c}{\textbf{Closed-source models}} \\
\midrule
\multirow{2}{*}{Offline}
& Gemini 1.5 Pro\cite{geminiteam2024gemini15unlockingmultimodal} & 69.32 & 62.54 & 57.15 & 63.00 & - & - & - \\
& GPT-4o\cite{openai2024gpt4ocard} & 64.46 & 60.75 & 53.40 & 59.54 & - & - & - \\
\midrule

\rowcolor{TabGroup}
\multicolumn{9}{c}{\textbf{Open-source models}} \\
\midrule
\multirow{4}{*}{Online}
& Flash-VStream-7B\cite{DBLP:journals/corr/abs-2506-23825} & 27.38 & 28.37 & 45.09 & 33.61 & - & - & - \\
& Dispider-7B\cite{DBLP:conf/cvpr/0001DD0ZCL025} & 36.06 & 54.55 & 34.72 & 41.78 & - & - & - \\
& StreamForest-7B\cite{DBLP:journals/corr/abs-2509-24871} & 52.02 & 61.20 & 53.49 & 55.57 & - & - & - \\
& StreamAgent-7B\cite{DBLP:journals/corr/abs-2508-01875} & 41.70 & 61.30 & 45.40 & 49.40 & - & - & - \\
\midrule

\rowcolor{TabGroup}
\multicolumn{9}{c}{\textbf{Qwen3-VL-2B}} \\
\midrule
\multirow{4}{*}{Offline}
& Instruct & 37.29 & 56.65 & \textbf{51.51} & 49.97 & +2.23 & - & - \\
& Thinking & 41.78 & 57.27 & 45.04 & 47.74 & +0.00 & 590.25 & 0.00 \\
& TWW$_{\text{Batch,S2}}$ & 43.25 & 56.52 & 44.72 & 47.67 & -0.07 & 518.64 & 12.13 \\
& \offc{\textbf{TWW$_{\text{Batch,S3}}$}}
& \offc{44.37}
& \offc{56.87}
& \offc{45.31}
& \offc{48.30}
& \offc{\textbf{+0.56}}
& \offc{530.32}
& \offc{10.15} \\
\midrule
\multirow{6}{*}{Online}
& Instruct$_{\text{online}}$ & 11.89 & 16.01 & 26.87 & 20.76 & -26.98 & - & - \\
& Thinking$_{\text{online}}$ & 8.56 & 19.47 & 20.29 & 17.63 & -30.11 & 478.74 & 18.89 \\
\cmidrule(lr){2-9}
& TWW$_{\text{single-turn,S2}}$ & 44.58 & 57.85 & 47.35 & 49.67 & +1.93 & 456.28 & 22.70 \\
& \sinc{\textbf{TWW$_{\text{single-turn,S3}}$}}
& \sinc{\textbf{45.96}}
& \sinc{\textbf{58.18}}
& \sinc{47.86}
& \sinc{\textbf{50.31}}
& \sinc{\textbf{+2.57}}
& \sinc{470.20}
& \sinc{20.34} \\
\cmidrule(lr){2-9}
& TWW$_{\text{multi-turn,S2}}$ & 40.85 & 55.42 & 44.34 & 46.67 & -1.07 & \textbf{278.52} & \textbf{52.81} \\
& \mulc{\textbf{TWW$_{\text{multi-turn,S3}}$}}
& \mulc{42.16}
& \mulc{55.79}
& \mulc{44.54}
& \mulc{47.15}
& \mulc{-0.59}
& \mulc{280.32}
& \mulc{52.51} \\
\midrule

\rowcolor{TabGroup}
\multicolumn{9}{c}{\textbf{Qwen3-VL-4B}} \\
\midrule
\multirow{4}{*}{Offline}
& Instruct & 44.32 & 62.39 & 46.29 & 50.32 & -0.38 & - & - \\
& Thinking & 50.78 & 61.17 & 45.08 & 50.70 & +0.00 & 472.18 & 0.00 \\
& TWW$_{\text{Batch,S2}}$ & 52.45 & 62.18 & 47.37 & 52.51 & +1.81 & 412.37 & 12.67 \\
& \offc{\textbf{TWW$_{\text{Batch,S3}}$}}
& \offc{53.88}
& \offc{62.37}
& \offc{47.86}
& \offc{53.11}
& \offc{\textbf{+2.41}}
& \offc{430.51}
& \offc{8.83} \\
\midrule
\multirow{6}{*}{Online}
& Instruct$_{\text{online}}$ & 13.63 & 20.79 & 24.95 & 21.45 & -29.25 & - & - \\
& Thinking$_{\text{online}}$ & 14.42 & 14.10 & 18.06 & 16.21 & -34.49 & 360.63 & 23.62 \\
\cmidrule(lr){2-9}
& TWW$_{\text{single-turn,S2}}$ & 53.92 & 64.25 & 49.55 & 54.51 & +3.81 & 358.45 & 24.09 \\
& \sinc{\textbf{TWW$_{\text{single-turn,S3}}$}}
& \sinc{\textbf{55.47}}
& \sinc{\textbf{64.52}}
& \sinc{\textbf{49.78}}
& \sinc{\textbf{55.02}}
& \sinc{\textbf{+4.32}}
& \sinc{378.64}
& \sinc{19.81} \\
\cmidrule(lr){2-9}
& TWW$_{\text{multi-turn,S2}}$ & 48.65 & 60.85 & 47.67 & 51.51 & +0.81 & \textbf{251.36} & \textbf{46.77} \\
& \mulc{\textbf{TWW$_{\text{multi-turn,S3}}$}}
& \mulc{49.13}
& \mulc{61.17}
& \mulc{47.86}
& \mulc{51.80}
& \mulc{+1.10}
& \mulc{255.91}
& \mulc{45.80} \\
\midrule

\rowcolor{TabGroup}
\multicolumn{9}{c}{\textbf{Qwen3-VL-8B}} \\
\midrule
\multirow{4}{*}{Offline}
& Instruct & 42.63 & 62.86 & 51.02 & 52.54 & -1.28 & - & - \\
& Thinking & 54.24 & 61.34 & 49.63 & 53.82 & +0.00 & 390.42 & 0.00 \\
& TWW$_{\text{Batch,S2}}$ & 55.82 & 63.15 & 51.83 & 55.78 & +1.96 & 325.82 & 16.55 \\
& \offc{\textbf{TWW$_{\text{Batch,S3}}$}}
& \offc{\textbf{57.37}}
& \offc{63.68}
& \offc{52.01}
& \offc{56.34}
& \offc{\textbf{+2.52}}
& \offc{340.22}
& \offc{12.86} \\
\midrule
\multirow{6}{*}{Online}
& Instruct$_{\text{online}}$ & 15.69 & 25.57 & 21.12 & 21.22 & -32.60 & - & - \\
& Thinking$_{\text{online}}$ & 11.25 & 19.59 & 18.51 & 17.30 & -36.52 & 330.67 & 15.30 \\
\cmidrule(lr){2-9}
& TWW$_{\text{single-turn,S2}}$ & 56.45 & 64.52 & 52.78 & 56.78 & +2.96 & 276.19 & 29.26 \\
& \sinc{\textbf{TWW$_{\text{single-turn,S3}}$}}
& \sinc{57.05}
& \sinc{\textbf{64.76}}
& \sinc{52.97}
& \sinc{\textbf{57.07}}
& \sinc{\textbf{+3.25}}
& \sinc{290.07}
& \sinc{25.70} \\
\cmidrule(lr){2-9}
& TWW$_{\text{multi-turn,S2}}$ & 52.05 & 63.25 & 52.85 & 55.55 & +1.73 & \textbf{224.68} & \textbf{42.45} \\
& \mulc{\textbf{TWW$_{\text{multi-turn,S3}}$}}
& \mulc{52.77}
& \mulc{63.68}
& \mulc{\textbf{53.29}}
& \mulc{56.05}
& \mulc{+2.23}
& \mulc{227.33}
& \mulc{41.77} \\
\bottomrule
\end{tabular}}
\end{table*}

\subsection{Analysis}
\label{sec:exp_analysis}

\begin{figure}[t]
  \centering
  \begin{minipage}[t]{0.49\linewidth}
    \centering
    \includegraphics[width=\linewidth]{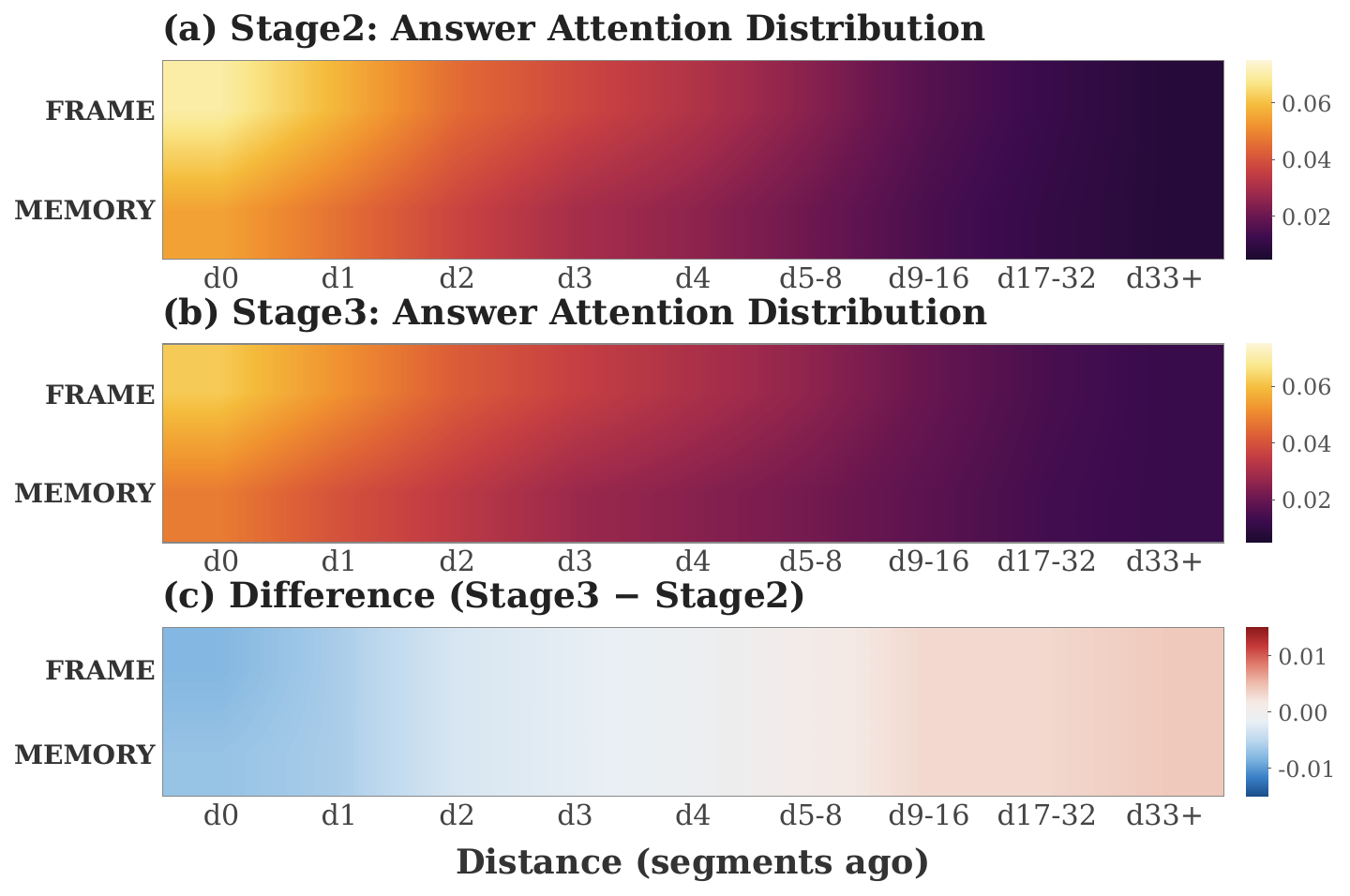}
    \caption{\textbf{Answer attention vs.\ segment distance} on TWW$_{\text{multi-turn}}$. After Stage~3, attention mass shifts from near-history to more distant segments.}
    \label{fig:attn_dist}
  \end{minipage}\hfill
  \begin{minipage}[t]{0.49\linewidth}
    \centering
    \includegraphics[width=\linewidth]{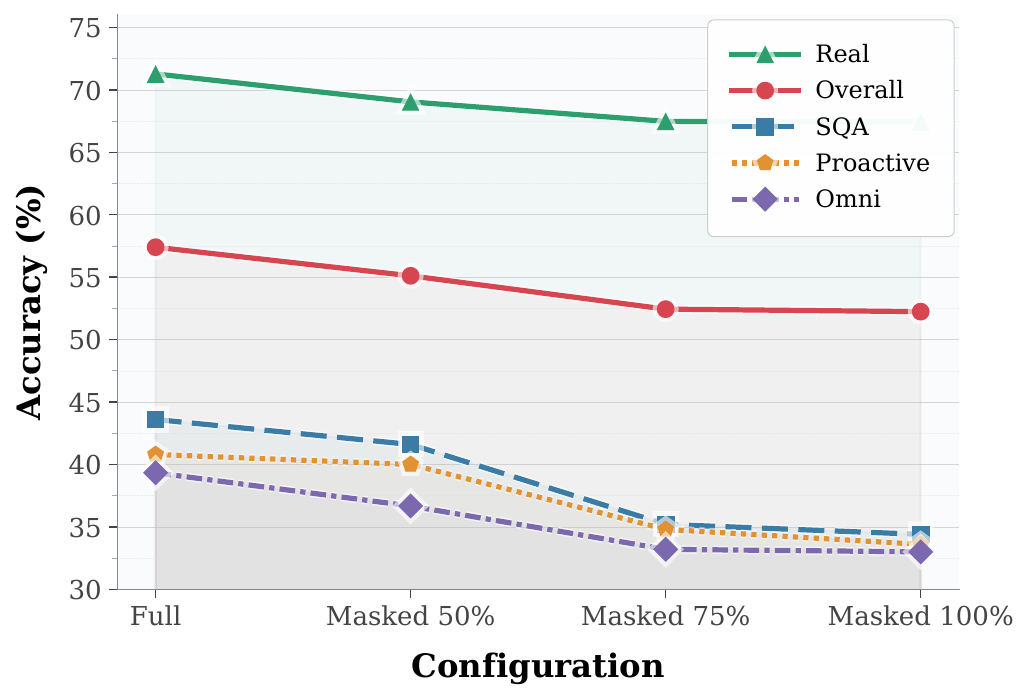}
    \caption{\textbf{Ablation} under frame masking on TWW$_{\text{multi-turn,S3}}$. Overall represents the accuracy rate. The remaining curves represent the results of the subsets.}
    \label{fig:frame_mask}
  \end{minipage}
\end{figure}

\refstepcounter{subsubsection}
\noindent\textbf{Generalization to Offline Video Understanding.}
\label{sec:offline_generalization}
Although our training is designed for streaming scenarios, we also evaluate whether the learned behaviors can transfer to offline video understanding tasks. We evaluate on two offline benchmarks: \textbf{Video-MME}~\cite{DBLP:conf/cvpr/FuDLLRZWZSZCLLZ25} and \textbf{LV-Bench}~\cite{wang2025lvbenchextremelongvideo}, following their official evaluation settings. \Cref{tab:offline_results} shows that streaming training also benefits offline evaluation. In particular, TWW$_{\text{single-turn,S3}}$ improves Video-MME from 68.89\% to \textbf{73.41\%} and LV-Bench from 53.47\% to \textbf{57.68\%}, showing that long-range streaming supervision transfers effectively to offline long-video reasoning. Unless otherwise stated, the following analyses use Qwen3-VL-4B on StreamingBench.

\begin{table*}[t]
\caption{\textbf{Offline video understanding results} on Video-MME and LV-Bench.}
\label{tab:offline_results}
\centering
\scriptsize
\setlength{\tabcolsep}{3.0pt}
\renewcommand{\arraystretch}{1.08}
\resizebox{\textwidth}{!}{
\begin{tabular}{@{}lccccccccccc@{}}
\toprule
& \multicolumn{4}{c}{\textbf{Video-MME}} & \multicolumn{7}{c}{\textbf{LV-Bench}} \\
\cmidrule(lr){2-5} \cmidrule(lr){6-12}
\textbf{Method} & \textbf{Short} & \textbf{Medium} & \textbf{Long} & \textbf{Overall} & \textbf{ER} & \textbf{EU} & \textbf{KIR} & \textbf{TG} & \textbf{Rea} & \textbf{Sum} & \textbf{Overall} \\
\midrule
Thinking & 77.78 & 66.67 & 62.22 & 68.89 & 54.22 & 51.62 & 62.07 & 46.82 & 55.50 & 40.35 & 53.47 \\
Instruct & 78.22 & 66.78 & 62.89 & 69.30 & 56.15 & 53.78 & 65.86 & 50.45 & 60.00 & 43.86 & 56.19 \\
TWW$_{\text{Batch,S3}}$ & 78.89 & 67.11 & 64.00 & 70.00 & \textbf{58.07} & 54.24 & 66.21 & \textbf{51.36} & 62.00 & 47.37 & 57.39 \\
TWW$_{\text{multi-turn,S3}}$ & 79.00 & 67.11 & 62.33 & 69.48 & 56.44 & 54.39 & 66.55 & 50.91 & 61.50 & 45.61 & 56.81 \\
TWW$_{\text{single-turn,S3}}$ & \textbf{83.22} & \textbf{70.11} & \textbf{66.89} & \textbf{73.41} & 57.33 & \textbf{54.70} & \textbf{67.59} & \textbf{51.36} & \textbf{62.50} & \textbf{52.63} & \textbf{57.68} \\
\bottomrule
\end{tabular}}
\end{table*}

\refstepcounter{subsubsection}
\noindent\textbf{Long-Range Attention Analysis.}
\label{sec:analysis_attention}
We analyze how far back the model consults history when generating answers by aggregating the last-layer attention from answer tokens to historical tokens and grouping the attended history by segment distance $d=\tau_r-i$, where $\tau_r$ denotes the index of the latest observed segment when answering the $r$-th question, and $i\le\tau_r$ denotes the index of a historical segment being attended to. Here, $d=0$ corresponds to the most recent segment, and larger $d$ indicates segments further in the past.

We separately measure attention to \textbf{FRAME} tokens, which are visual tokens from prior segments, and \textbf{MEMORY} tokens, which are accumulated memory note tokens written for those segments. \Cref{fig:attn_dist} shows that the Stage~2 checkpoint exhibits a strong recency bias, whereas Stage~3 reallocates attention mass from near-history buckets to more distant buckets. The shift is more pronounced on \textbf{MEMORY} tokens than on visual tokens, consistent with the intended role of memory notes as a compact long-range state for multi-turn interaction.

\refstepcounter{subsubsection}
\noindent\textbf{Ablation Studies.}
\label{sec:analysis_ablation}
We conduct ablations to isolate the roles of the memory bank, visual inputs, and segmentation granularity. Removing the memory bank causes a clear accuracy drop from $57.40$\% to $52.35$\% in \Cref{tab:memory_ablation}, confirming that memory notes serve as an effective persistent state in multi-round streams. For visual ablations, Fig.~\ref{fig:frame_mask} shows a monotonic degradation as more frames are masked. Performance remains relatively stable under moderate corruption, suggesting that once written, segment-level memory notes provide a stabilizing signal. Under severe corruption, accuracy approaches the no-memory regime, which is expected because memory writing becomes unreliable without sufficient visual evidence. For segmentation granularity, Table~\ref{tab:memory_ablation} reveals a clear accuracy and efficiency tradeoff. Using longer segments $120$s/$60$s reduces the average decoding length from 302.56 to 230.46 tokens but causes a noticeable accuracy drop of $2.07$\%. Conversely, using shorter segments $30$s/$15$s preserves accuracy but increases the average decoding length to 380.50 tokens (+$25.8$\%), due to more frequent memory updates.

\begin{table}[t]
\centering
\begin{minipage}[t]{0.49\linewidth}
\centering
\captionof{table}{\textbf{Ablations} on StreamingBench. a/b represents the maximum segment duration and the chunk duration.}
\label{tab:memory_ablation}
{\scriptsize
\setlength{\tabcolsep}{4pt}
\renewcommand{\arraystretch}{1.06}
\begin{tabular}{@{}llcc@{}}
\toprule
\textbf{Category} & \textbf{Setting} & \textbf{Acc}$\uparrow$ & \textbf{Avg Tok}$\downarrow$ \\
\midrule
Memory & with notes & \textbf{57.40} & 302.56 \\
Memory & without notes & 52.35 & 330.73 \\
\midrule
Segment & $60$s/$30$s & \textbf{57.40} & 302.56 \\
Segment & $120$s/$60$s & 55.33 & \textbf{230.46} \\
Segment & $30$s/$15$s & 57.20 & 380.50 \\
\bottomrule
\end{tabular}}
\end{minipage}\hfill
\begin{minipage}[t]{0.49\linewidth}
\centering
\captionof{table}{\textbf{TTFT on StreamingBench.} Overall accuracy and time-to-first-token of Qwen3-VL-4B for batch, interleaved streaming, and our TWW$_{\text{multi-turn,S3}}$ pipeline.}
\label{tab:ttft}
{\scriptsize
\setlength{\tabcolsep}{6pt}
\renewcommand{\arraystretch}{1.06}
\begin{tabular}{@{}lcc@{}}
\toprule
\textbf{Method} & \textbf{Overall Acc}$\uparrow$ & \textbf{TTFT}$\downarrow$ \\
\midrule
Thinking & \textbf{58.52} & 31203.69 \\
Interleaved & 55.35 & \textbf{2304.28} \\
TWW$_{\text{multi-turn,S3}}$ & 57.40 & \textbf{2304.28} \\
\bottomrule
\end{tabular}}
\end{minipage}
\end{table}

\refstepcounter{subsubsection}
\noindent\textbf{TTFT Analysis.}
\label{sec:analysis_ttft}
Table~\ref{tab:ttft} reports overall accuracy and TTFT on StreamingBench with the 4B backbone. Compared with batch Thinking, our streaming pipeline reduces TTFT by \textbf{92.6\%}, from 31203.69 to 2304.28 tokens, while maintaining comparable accuracy. The interleaved baseline achieves a similar TTFT for multi-turn streaming but is consistently less accurate.

\refstepcounter{subsubsection}
\noindent\textbf{Theoretical Latency Analysis.}
\label{sec:analysis_theory_latency}
Our method decouples ingestion from decoding, largely eliminating decoder-induced ingestion backlog, thereby avoiding backlog explosion as $\rho$ (arrival rate over processing rate) approaches $1$ and significantly reducing latency. The complete derivation is in Appendix~\ref{app:latency}.

\section{Conclusion}

We presented \textbf{Think While Watching}, a memory-anchored streaming video reasoning framework for multi-turn interaction over continuously arriving streams. Our approach maintains \textbf{segment-level memory notes} as a persistent state, enforces strict causality through a \textbf{segment-level streaming causal mask} and \textbf{streaming positional encoding}, and enables practical real-time deployment via a dual KV cache pipeline with adaptive attention backends. Experiments on StreamingBench and OVO-Bench validate the proposed method’s effectiveness, consistently improving online accuracy while maintaining strong efficiency.

\bibliographystyle{unsrtnat}
\bibliography{main}

@String(CVPR  = {IEEE Conf. Comput. Vis. Pattern Recog.})

@String(ICLR  = {Int. Conf. Learn. Represent.})

@String(CVPR  = {CVPR})

@String(ICLR  = {ICLR})

@misc{lin2024streamingbenchassessinggapmllms,
      title={StreamingBench: Assessing the Gap for MLLMs to Achieve Streaming Video Understanding}, 
      author={Junming Lin and Zheng Fang and Chi Chen and Zihao Wan and Fuwen Luo and Peng Li and Yang Liu and Maosong Sun},
      year={2024},
      eprint={2411.03628},
      archivePrefix={arXiv},
      primaryClass={cs.CV},
      url={https://arxiv.org/abs/2411.03628}, 
}

@article{Yang2025ViSPeak,
  author       = {Shenghao Fu and
                  Qize Yang and
                  Yuan{-}Ming Li and
                  Yi{-}Xing Peng and
                  Kun{-}Yu Lin and
                  Xihan Wei and
                  Jian{-}Fang Hu and
                  Xiaohua Xie and
                  Wei{-}Shi Zheng},
  title        = {ViSpeak: Visual Instruction Feedback in Streaming Videos},
  journal      = {CoRR},
  volume       = {abs/2503.12769},
  year         = {2025},
  url          = {https://doi.org/10.48550/arXiv.2503.12769},
  doi          = {10.48550/ARXIV.2503.12769},
  eprinttype    = {arXiv},
  eprint       = {2503.12769},
  bibsource    = {dblp computer science bibliography, https://dblp.org}
}

@article{DBLP:journals/corr/abs-2510-25332,
  author       = {Yuhang Hu and
                  Zhenyu Yang and
                  Shihan Wang and
                  Shengsheng Qian and
                  Bin Wen and
                  Fan Yang and
                  Tingting Gao and
                  Changsheng Xu},
  title        = {StreamingCoT: {A} Dataset for Temporal Dynamics and Multimodal Chain-of-Thought
                  Reasoning in Streaming VideoQA},
  journal      = {CoRR},
  volume       = {abs/2510.25332},
  year         = {2025},
  url          = {https://doi.org/10.48550/arXiv.2510.25332},
  doi          = {10.48550/ARXIV.2510.25332},
  eprinttype    = {arXiv},
  eprint       = {2510.25332},
  bibsource    = {dblp computer science bibliography, https://dblp.org}
}

@article{DBLP:journals/corr/abs-2505-02064,
  author       = {Shuhang Xun and
                  Sicheng Tao and
                  Jungang Li and
                  Yibo Shi and
                  Zhixin Lin and
                  Zhanhui Zhu and
                  Yibo Yan and
                  Hanqian Li and
                  Linghao Zhang and
                  Shikang Wang and
                  Yixin Liu and
                  Hanbo Zhang and
                  Ying Ma and
                  Xuming Hu},
  title        = {RTV-Bench: Benchmarking {MLLM} Continuous Perception, Understanding
                  and Reasoning through Real-Time Video},
  journal      = {CoRR},
  volume       = {abs/2505.02064},
  year         = {2025},
  url          = {https://doi.org/10.48550/arXiv.2505.02064},
  doi          = {10.48550/ARXIV.2505.02064},
  eprinttype    = {arXiv},
  eprint       = {2505.02064},
  bibsource    = {dblp computer science bibliography, https://dblp.org}
}

@inproceedings{DBLP:conf/cvpr/NiuLMGZHDDD0ZZC25,
  author       = {Junbo Niu and
                  Yifei Li and
                  Ziyang Miao and
                  Chunjiang Ge and
                  Yuanhang Zhou and
                  Qihao He and
                  Xiaoyi Dong and
                  Haodong Duan and
                  Shuangrui Ding and
                  Rui Qian and
                  Pan Zhang and
                  Yuhang Zang and
                  Yuhang Cao and
                  Conghui He and
                  Jiaqi Wang},
  title        = {OVO-Bench: How Far is Your Video-LLMs from Real-World Online Video
                  Understanding?},
  booktitle    = {{IEEE/CVF} Conference on Computer Vision and Pattern Recognition,
                  {CVPR} 2025, Nashville, TN, USA, June 11-15, 2025},
  pages        = {18902--18913},
  publisher    = {Computer Vision Foundation / {IEEE}},
  year         = {2025},
  url          = {https://openaccess.thecvf.com/content/CVPR2025/html/Niu\_OVO-Bench\_How\_Far\_is\_Your\_Video-LLMs\_from\_Real-World\_Online\_Video\_CVPR\_2025\_paper.html},
  doi          = {10.1109/CVPR52734.2025.01761},
  bibsource    = {dblp computer science bibliography, https://dblp.org}
}

@misc{lee2025streamgazegazeguidedtemporalreasoning,
      title={StreamGaze: Gaze-Guided Temporal Reasoning and Proactive Understanding in Streaming Videos}, 
      author={Daeun Lee and Subhojyoti Mukherjee and Branislav Kveton and Ryan A. Rossi and Viet Dac Lai and Seunghyun Yoon and Trung Bui and Franck Dernoncourt and Mohit Bansal},
      year={2025},
      eprint={2512.01707},
      archivePrefix={arXiv},
      primaryClass={cs.CV},
      url={https://arxiv.org/abs/2512.01707}, 
}

@inproceedings{DBLP:conf/iclr/YangHDXQW0DX25,
  author       = {Zhenyu Yang and
                  Yuhang Hu and
                  Zemin Du and
                  Dizhan Xue and
                  Shengsheng Qian and
                  Jiahong Wu and
                  Fan Yang and
                  Weiming Dong and
                  Changsheng Xu},
  title        = {SVBench: {A} Benchmark with Temporal Multi-Turn Dialogues for Streaming
                  Video Understanding},
  booktitle    = {The Thirteenth International Conference on Learning Representations,
                  {ICLR} 2025, Singapore, April 24-28, 2025},
  publisher    = {OpenReview.net},
  year         = {2025},
  url          = {https://openreview.net/forum?id=Hz4BYVY8YM},
  bibsource    = {dblp computer science bibliography, https://dblp.org}
}

@misc{wang2025streameqastreamingvideounderstanding,
      title={StreamEQA: Towards Streaming Video Understanding for Embodied Scenarios}, 
      author={Yifei Wang and Zhenkai Li and Tianwen Qian and Huanran Zheng and Zheng Wang and Yuqian Fu and Xiaoling Wang},
      year={2025},
      eprint={2512.04451},
      archivePrefix={arXiv},
      primaryClass={cs.CV},
      url={https://arxiv.org/abs/2512.04451}, 
}

@inproceedings{DBLP:conf/cvpr/ChenLWLSGLGMS24,
  author       = {Joya Chen and
                  Zhaoyang Lv and
                  Shiwei Wu and
                  Kevin Qinghong Lin and
                  Chenan Song and
                  Difei Gao and
                  Jia{-}Wei Liu and
                  Ziteng Gao and
                  Dongxing Mao and
                  Mike Zheng Shou},
  title        = {VideoLLM-online: Online Video Large Language Model for Streaming Video},
  booktitle    = {{IEEE/CVF} Conference on Computer Vision and Pattern Recognition,
                  {CVPR} 2024, Seattle, WA, USA, June 16-22, 2024},
  pages        = {18407--18418},
  publisher    = {{IEEE}},
  year         = {2024},
  url          = {https://doi.org/10.1109/CVPR52733.2024.01742},
  doi          = {10.1109/CVPR52733.2024.01742},
  bibsource    = {dblp computer science bibliography, https://dblp.org}
}

@misc{kim2025vrexrealtimestreamingvideo,
      title={V-Rex: Real-Time Streaming Video LLM Acceleration via Dynamic KV Cache Retrieval}, 
      author={Donghyuk Kim and Sejeong Yang and Wonjin Shin and Joo-Young Kim},
      year={2025},
      eprint={2512.12284},
      archivePrefix={arXiv},
      primaryClass={eess.IV},
      url={https://arxiv.org/abs/2512.12284}, 
}

@article{DBLP:journals/corr/abs-2504-17343,
  author       = {Linli Yao and
                  Yicheng Li and
                  Yuancheng Wei and
                  Lei Li and
                  Shuhuai Ren and
                  Yuanxin Liu and
                  Kun Ouyang and
                  Lean Wang and
                  Shicheng Li and
                  Sida Li and
                  Lingpeng Kong and
                  Qi Liu and
                  Yuanxing Zhang and
                  Xu Sun},
  title        = {TimeChat-Online: 80{\%} Visual Tokens are Naturally Redundant in Streaming
                  Videos},
  journal      = {CoRR},
  volume       = {abs/2504.17343},
  year         = {2025},
  url          = {https://doi.org/10.48550/arXiv.2504.17343},
  doi          = {10.48550/ARXIV.2504.17343},
  eprinttype    = {arXiv},
  eprint       = {2504.17343},
  bibsource    = {dblp computer science bibliography, https://dblp.org}
}

@article{DBLP:journals/corr/abs-2511-07278,
  author       = {Yilong Chen and
                  Xiang Bai and
                  Zhibin Wang and
                  Chengyu Bai and
                  Yuhan Dai and
                  Ming Lu and
                  Shanghang Zhang},
  title        = {StreamKV: Streaming Video Question-Answering with Segment-based {KV}
                  Cache Retrieval and Compression},
  journal      = {CoRR},
  volume       = {abs/2511.07278},
  year         = {2025},
  url          = {https://doi.org/10.48550/arXiv.2511.07278},
  doi          = {10.48550/ARXIV.2511.07278},
  eprinttype    = {arXiv},
  eprint       = {2511.07278},
  bibsource    = {dblp computer science bibliography, https://dblp.org}
}

@article{DBLP:journals/corr/abs-2510-09608,
  author       = {Ruyi Xu and
                  Guangxuan Xiao and
                  Yukang Chen and
                  Liuning He and
                  Kelly Peng and
                  Yao Lu and
                  Song Han},
  title        = {StreamingVLM: Real-Time Understanding for Infinite Video Streams},
  journal      = {CoRR},
  volume       = {abs/2510.09608},
  year         = {2025},
  url          = {https://doi.org/10.48550/arXiv.2510.09608},
  doi          = {10.48550/ARXIV.2510.09608},
  eprinttype    = {arXiv},
  eprint       = {2510.09608},
  bibsource    = {dblp computer science bibliography, https://dblp.org}
}

@article{DBLP:journals/corr/abs-2510-18269,
  author       = {Xueyi Chen and
                  Keda Tao and
                  Kele Shao and
                  Huan Wang},
  title        = {StreamingTOM: Streaming Token Compression for Efficient Video Understanding},
  journal      = {CoRR},
  volume       = {abs/2510.18269},
  year         = {2025},
  url          = {https://doi.org/10.48550/arXiv.2510.18269},
  doi          = {10.48550/ARXIV.2510.18269},
  eprinttype    = {arXiv},
  eprint       = {2510.18269},
  bibsource    = {dblp computer science bibliography, https://dblp.org}
}

@article{DBLP:journals/corr/abs-2510-17238,
  author       = {Junlong Tong and
                  Yingqi Fan and
                  Anhao Zhao and
                  Yunpu Ma and
                  Xiaoyu Shen},
  title        = {StreamingThinker: Large Language Models Can Think While Reading},
  journal      = {CoRR},
  volume       = {abs/2510.17238},
  year         = {2025},
  url          = {https://doi.org/10.48550/arXiv.2510.17238},
  doi          = {10.48550/ARXIV.2510.17238},
  eprinttype    = {arXiv},
  eprint       = {2510.17238},
  bibsource    = {dblp computer science bibliography, https://dblp.org}
}

@inproceedings{DBLP:conf/iclr/XiongYYZ0ZL25,
  author       = {Haomiao Xiong and
                  Zongxin Yang and
                  Jiazuo Yu and
                  Yunzhi Zhuge and
                  Lu Zhang and
                  Jiawen Zhu and
                  Huchuan Lu},
  title        = {Streaming Video Understanding and Multi-round Interaction with Memory-enhanced
                  Knowledge},
  booktitle    = {The Thirteenth International Conference on Learning Representations,
                  {ICLR} 2025, Singapore, April 24-28, 2025},
  publisher    = {OpenReview.net},
  year         = {2025},
  url          = {https://openreview.net/forum?id=JbPb6RieNC},
  bibsource    = {dblp computer science bibliography, https://dblp.org}
}

@inproceedings{DBLP:conf/iclr/DiYZLZCLHSJ25,
  author       = {Shangzhe Di and
                  Zhelun Yu and
                  Guanghao Zhang and
                  Haoyuan Li and
                  Tao Zhong and
                  Hao Cheng and
                  Bolin Li and
                  Wanggui He and
                  Fangxun Shu and
                  Hao Jiang},
  title        = {Streaming Video Question-Answering with In-context Video KV-Cache
                  Retrieval},
  booktitle    = {The Thirteenth International Conference on Learning Representations,
                  {ICLR} 2025, Singapore, April 24-28, 2025},
  publisher    = {OpenReview.net},
  year         = {2025},
  url          = {https://openreview.net/forum?id=8g9fs6mdEG},
  bibsource    = {dblp computer science bibliography, https://dblp.org}
}

@inproceedings{DBLP:conf/cvpr/ZhouABYMXNS24,
  author       = {Xingyi Zhou and
                  Anurag Arnab and
                  Shyamal Buch and
                  Shen Yan and
                  Austin Myers and
                  Xuehan Xiong and
                  Arsha Nagrani and
                  Cordelia Schmid},
  title        = {Streaming Dense Video Captioning},
  booktitle    = {{IEEE/CVF} Conference on Computer Vision and Pattern Recognition,
                  {CVPR} 2024, Seattle, WA, USA, June 16-22, 2024},
  pages        = {18243--18252},
  publisher    = {{IEEE}},
  year         = {2024},
  url          = {https://doi.org/10.1109/CVPR52733.2024.01727},
  doi          = {10.1109/CVPR52733.2024.01727},
  bibsource    = {dblp computer science bibliography, https://dblp.org}
}

@article{DBLP:journals/corr/abs-2509-24871,
  author       = {Xiangyu Zeng and
                  Kefan Qiu and
                  Qingyu Zhang and
                  Xinhao Li and
                  Jing Wang and
                  Jiaxin Li and
                  Ziang Yan and
                  Kun Tian and
                  Meng Tian and
                  Xinhai Zhao and
                  Yi Wang and
                  Limin Wang},
  title        = {StreamForest: Efficient Online Video Understanding with Persistent
                  Event Memory},
  journal      = {CoRR},
  volume       = {abs/2509.24871},
  year         = {2025},
  url          = {https://doi.org/10.48550/arXiv.2509.24871},
  doi          = {10.48550/ARXIV.2509.24871},
  eprinttype    = {arXiv},
  eprint       = {2509.24871},
  bibsource    = {dblp computer science bibliography, https://dblp.org}
}

@article{DBLP:journals/corr/abs-2412-08646,
  author       = {Jihao Liu and
                  Zhiding Yu and
                  Shiyi Lan and
                  Shihao Wang and
                  Rongyao Fang and
                  Jan Kautz and
                  Hongsheng Li and
                  Jos{\'{e}} M. {\'{A}}lvarez},
  title        = {StreamChat: Chatting with Streaming Video},
  journal      = {CoRR},
  volume       = {abs/2412.08646},
  year         = {2024},
  url          = {https://doi.org/10.48550/arXiv.2412.08646},
  doi          = {10.48550/ARXIV.2412.08646},
  eprinttype    = {arXiv},
  eprint       = {2412.08646},
  bibsource    = {dblp computer science bibliography, https://dblp.org}
}

@article{DBLP:journals/corr/abs-2508-01875,
  author       = {Haolin Yang and
                  Feilong Tang and
                  Linxiao Zhao and
                  Xiang An and
                  Ming Hu and
                  Huifa Li and
                  Xinlin Zhuang and
                  Boqian Wang and
                  Yifan Lu and
                  Xiaofeng Zhang and
                  Abdalla Swikir and
                  Junjun He and
                  Zongyuan Ge and
                  Imran Razzak},
  title        = {StreamAgent: Towards Anticipatory Agents for Streaming Video Understanding},
  journal      = {CoRR},
  volume       = {abs/2508.01875},
  year         = {2025},
  url          = {https://doi.org/10.48550/arXiv.2508.01875},
  doi          = {10.48550/ARXIV.2508.01875},
  eprinttype    = {arXiv},
  eprint       = {2508.01875},
  bibsource    = {dblp computer science bibliography, https://dblp.org}
}

@inproceedings{DBLP:conf/acl/TongFLFZSS25,
  author       = {Junlong Tong and
                  Jinlan Fu and
                  Zixuan Lin and
                  Yingqi Fan and
                  Anhao Zhao and
                  Hui Su and
                  Xiaoyu Shen},
  editor       = {Wanxiang Che and
                  Joyce Nabende and
                  Ekaterina Shutova and
                  Mohammad Taher Pilehvar},
  title        = {{LLM} as Effective Streaming Processor: Bridging Streaming-Batch Mismatches
                  with Group Position Encoding},
  booktitle    = {Findings of the Association for Computational Linguistics, {ACL} 2025,
                  Vienna, Austria, July 27 - August 1, 2025},
  pages        = {23497--23517},
  publisher    = {Association for Computational Linguistics},
  year         = {2025},
  url          = {https://aclanthology.org/2025.findings-acl.1207/},
  bibsource    = {dblp computer science bibliography, https://dblp.org}
}

@article{DBLP:journals/corr/abs-2505-15269,
  author       = {Zhenyu Ning and
                  Guangda Liu and
                  Qihao Jin and
                  Wenchao Ding and
                  Minyi Guo and
                  Jieru Zhao},
  title        = {LiveVLM: Efficient Online Video Understanding via Streaming-Oriented
                  {KV} Cache and Retrieval},
  journal      = {CoRR},
  volume       = {abs/2505.15269},
  year         = {2025},
  url          = {https://doi.org/10.48550/arXiv.2505.15269},
  doi          = {10.48550/ARXIV.2505.15269},
  eprinttype    = {arXiv},
  eprint       = {2505.15269},
  bibsource    = {dblp computer science bibliography, https://dblp.org}
}

@article{DBLP:journals/corr/abs-2511-05299,
  author       = {Zhenyu Yang and
                  Kairui Zhang and
                  Yuhang Hu and
                  Bing Wang and
                  Shengsheng Qian and
                  Bin Wen and
                  Fan Yang and
                  Tingting Gao and
                  Weiming Dong and
                  Changsheng Xu},
  title        = {LiveStar: Live Streaming Assistant for Real-World Online Video Understanding},
  journal      = {CoRR},
  volume       = {abs/2511.05299},
  year         = {2025},
  url          = {https://doi.org/10.48550/arXiv.2511.05299},
  doi          = {10.48550/ARXIV.2511.05299},
  eprinttype    = {arXiv},
  eprint       = {2511.05299},
  bibsource    = {dblp computer science bibliography, https://dblp.org}
}

@inproceedings{DBLP:conf/cvpr/ChenZLLMS25,
  author       = {Joya Chen and
                  Ziyun Zeng and
                  Yiqi Lin and
                  Wei Li and
                  Zejun Ma and
                  Mike Zheng Shou},
  title        = {LiveCC: Learning Video {LLM} with Streaming Speech Transcription at
                  Scale},
  booktitle    = {{IEEE/CVF} Conference on Computer Vision and Pattern Recognition,
                  {CVPR} 2025, Nashville, TN, USA, June 11-15, 2025},
  pages        = {29083--29095},
  publisher    = {Computer Vision Foundation / {IEEE}},
  year         = {2025},
  url          = {https://openaccess.thecvf.com/content/CVPR2025/html/Chen\_LiveCC\_Learning\_Video\_LLM\_with\_Streaming\_Speech\_Transcription\_at\_Scale\_CVPR\_2025\_paper.html},
  doi          = {10.1109/CVPR52734.2025.02708},
  bibsource    = {dblp computer science bibliography, https://dblp.org}
}

@article{DBLP:journals/corr/abs-2506-23825,
  author       = {Haoji Zhang and
                  Yiqin Wang and
                  Yansong Tang and
                  Yong Liu and
                  Jiashi Feng and
                  Xiaojie Jin},
  title        = {Flash-VStream: Efficient Real-Time Understanding for Long Video Streams},
  journal      = {CoRR},
  volume       = {abs/2506.23825},
  year         = {2025},
  url          = {https://doi.org/10.48550/arXiv.2506.23825},
  doi          = {10.48550/ARXIV.2506.23825},
  eprinttype    = {arXiv},
  eprint       = {2506.23825},
  bibsource    = {dblp computer science bibliography, https://dblp.org}
}

@article{DBLP:journals/corr/abs-2510-14560,
  author       = {Yulin Zhang and
                  Cheng Shi and
                  Yang Wang and
                  Sibei Yang},
  title        = {Eyes Wide Open: Ego Proactive Video-LLM for Streaming Video},
  journal      = {CoRR},
  volume       = {abs/2510.14560},
  year         = {2025},
  url          = {https://doi.org/10.48550/arXiv.2510.14560},
  doi          = {10.48550/ARXIV.2510.14560},
  eprinttype    = {arXiv},
  eprint       = {2510.14560},
  bibsource    = {dblp computer science bibliography, https://dblp.org}
}

@article{DBLP:journals/corr/abs-2505-13140,
  author       = {Takahiro Maeda and
                  Jinkun Cao and
                  Norimichi Ukita and
                  Kris Kitani},
  title        = {CacheFlow: Fast Human Motion Prediction by Cached Normalizing Flow},
  journal      = {CoRR},
  volume       = {abs/2505.13140},
  year         = {2025},
  url          = {https://doi.org/10.48550/arXiv.2505.13140},
  doi          = {10.48550/ARXIV.2505.13140},
  eprinttype    = {arXiv},
  eprint       = {2505.13140},
  bibsource    = {dblp computer science bibliography, https://dblp.org}
}

@misc{lin2026speakwatchingunleashingtrue,
      title={Speak While Watching: Unleashing TRUE Real-Time Video Understanding Capability of Multimodal Large Language Models}, 
      author={Junyan Lin and Junlong Tong and Hao Wu and Jialiang Zhang and Jinming Liu and Xin Jin and Xiaoyu Shen},
      year={2026},
      eprint={2601.06843},
      archivePrefix={arXiv},
      primaryClass={cs.CV},
      url={https://arxiv.org/abs/2601.06843}, 
}

@article{DBLP:journals/corr/abs-2508-09736,
  author       = {Lin Long and
                  Yichen He and
                  Wentao Ye and
                  Yiyuan Pan and
                  Yuan Lin and
                  Hang Li and
                  Junbo Zhao and
                  Wei Li},
  title        = {Seeing, Listening, Remembering, and Reasoning: {A} Multimodal Agent
                  with Long-Term Memory},
  journal      = {CoRR},
  volume       = {abs/2508.09736},
  year         = {2025},
  url          = {https://doi.org/10.48550/arXiv.2508.09736},
  doi          = {10.48550/ARXIV.2508.09736},
  eprinttype    = {arXiv},
  eprint       = {2508.09736},
  bibsource    = {dblp computer science bibliography, https://dblp.org}
}

@article{DBLP:journals/corr/abs-2501-03230,
  author       = {Hao Fei and
                  Shengqiong Wu and
                  Wei Ji and
                  Hanwang Zhang and
                  Meishan Zhang and
                  Mong{-}Li Lee and
                  Wynne Hsu},
  title        = {Video-of-Thought: Step-by-Step Video Reasoning from Perception to
                  Cognition},
  journal      = {CoRR},
  volume       = {abs/2501.03230},
  year         = {2025},
  url          = {https://doi.org/10.48550/arXiv.2501.03230},
  doi          = {10.48550/ARXIV.2501.03230},
  eprinttype    = {arXiv},
  eprint       = {2501.03230},
  bibsource    = {dblp computer science bibliography, https://dblp.org}
}

@article{DBLP:journals/corr/abs-2503-21776,
  author       = {Kaituo Feng and
                  Kaixiong Gong and
                  Bohao Li and
                  Zonghao Guo and
                  Yibing Wang and
                  Tianshuo Peng and
                  Benyou Wang and
                  Xiangyu Yue},
  title        = {Video-R1: Reinforcing Video Reasoning in MLLMs},
  journal      = {CoRR},
  volume       = {abs/2503.21776},
  year         = {2025},
  url          = {https://doi.org/10.48550/arXiv.2503.21776},
  doi          = {10.48550/ARXIV.2503.21776},
  eprinttype    = {arXiv},
  eprint       = {2503.21776},
  bibsource    = {dblp computer science bibliography, https://dblp.org}
}

@article{DBLP:journals/corr/abs-2508-04416,
  author       = {Haoji Zhang and
                  Xin Gu and
                  Jiawen Li and
                  Chixiang Ma and
                  Sule Bai and
                  Chubin Zhang and
                  Bowen Zhang and
                  Zhichao Zhou and
                  Dongliang He and
                  Yansong Tang},
  title        = {Thinking With Videos: Multimodal Tool-Augmented Reinforcement Learning
                  for Long Video Reasoning},
  journal      = {CoRR},
  volume       = {abs/2508.04416},
  year         = {2025},
  url          = {https://doi.org/10.48550/arXiv.2508.04416},
  doi          = {10.48550/ARXIV.2508.04416},
  eprinttype    = {arXiv},
  eprint       = {2508.04416},
  bibsource    = {dblp computer science bibliography, https://dblp.org}
}

@article{DBLP:journals/corr/abs-2511-04570,
  author       = {Jingqi Tong and
                  Yurong Mou and
                  Hangcheng Li and
                  Mingzhe Li and
                  Yongzhuo Yang and
                  Ming Zhang and
                  Qiguang Chen and
                  Tianyi Liang and
                  Xiaomeng Hu and
                  Yining Zheng and
                  Xinchi Chen and
                  Jun Zhao and
                  Xuanjing Huang and
                  Xipeng Qiu},
  title        = {Thinking with Video: Video Generation as a Promising Multimodal Reasoning
                  Paradigm},
  journal      = {CoRR},
  volume       = {abs/2511.04570},
  year         = {2025},
  url          = {https://doi.org/10.48550/arXiv.2511.04570},
  doi          = {10.48550/ARXIV.2511.04570},
  eprinttype    = {arXiv},
  eprint       = {2511.04570},
  bibsource    = {dblp computer science bibliography, https://dblp.org}
}

@book{DBLP:books/wi/SGG2018,
  author       = {Abraham Silberschatz and
                  Peter Baer Galvin and
                  Greg Gagne},
  title        = {Operating System Concepts, 10th Edition},
  publisher    = {Wiley},
  year         = {2018},
  url          = {http://os-book.com/OS10/index.html},
  isbn         = {978-1-118-06333-0},
  bibsource    = {dblp computer science bibliography, https://dblp.org}
}

@article{DBLP:journals/corr/abs-2507-07966,
  author       = {Yukang Chen and
                  Wei Huang and
                  Baifeng Shi and
                  Qinghao Hu and
                  Hanrong Ye and
                  Ligeng Zhu and
                  Zhijian Liu and
                  Pavlo Molchanov and
                  Jan Kautz and
                  Xiaojuan Qi and
                  Sifei Liu and
                  Hongxu Yin and
                  Yao Lu and
                  Song Han},
  title        = {Scaling {RL} to Long Videos},
  journal      = {CoRR},
  volume       = {abs/2507.07966},
  year         = {2025},
  url          = {https://doi.org/10.48550/arXiv.2507.07966},
  doi          = {10.48550/ARXIV.2507.07966},
  eprinttype    = {arXiv},
  eprint       = {2507.07966},
  bibsource    = {dblp computer science bibliography, https://dblp.org}
}

@article{DBLP:journals/corr/abs-2409-12191,
  author       = {Peng Wang and
                  Shuai Bai and
                  Sinan Tan and
                  Shijie Wang and
                  Zhihao Fan and
                  Jinze Bai and
                  Keqin Chen and
                  Xuejing Liu and
                  Jialin Wang and
                  Wenbin Ge and
                  Yang Fan and
                  Kai Dang and
                  Mengfei Du and
                  Xuancheng Ren and
                  Rui Men and
                  Dayiheng Liu and
                  Chang Zhou and
                  Jingren Zhou and
                  Junyang Lin},
  title        = {Qwen2-VL: Enhancing Vision-Language Model's Perception of the
                  World at Any Resolution},
  journal      = {CoRR},
  volume       = {abs/2409.12191},
  year         = {2024},
  url          = {https://doi.org/10.48550/arXiv.2409.12191},
  doi          = {10.48550/ARXIV.2409.12191},
  eprinttype    = {arXiv},
  eprint       = {2409.12191},
  bibsource    = {dblp computer science bibliography, https://dblp.org}
}

@inproceedings{DBLP:conf/cvpr/FuDLLRZWZSZCLLZ25,
  author       = {Chaoyou Fu and
                  Yuhan Dai and
                  Yongdong Luo and
                  Lei Li and
                  Shuhuai Ren and
                  Renrui Zhang and
                  Zihan Wang and
                  Chenyu Zhou and
                  Yunhang Shen and
                  Mengdan Zhang and
                  Peixian Chen and
                  Yanwei Li and
                  Shaohui Lin and
                  Sirui Zhao and
                  Ke Li and
                  Tong Xu and
                  Xiawu Zheng and
                  Enhong Chen and
                  Caifeng Shan and
                  Ran He and
                  Xing Sun},
  title        = {Video-MME: The First-Ever Comprehensive Evaluation Benchmark of Multi-modal
                  LLMs in Video Analysis},
  booktitle    = {{IEEE/CVF} Conference on Computer Vision and Pattern Recognition,
                  {CVPR} 2025, Nashville, TN, USA, June 11-15, 2025},
  pages        = {24108--24118},
  publisher    = {Computer Vision Foundation / {IEEE}},
  year         = {2025},
  url          = {https://openaccess.thecvf.com/content/CVPR2025/html/Fu\_Video-MME\_The\_First-Ever\_Comprehensive\_Evaluation\_Benchmark\_of\_Multi-modal\_LLMs\_in\_CVPR\_2025\_paper.html},
  doi          = {10.1109/CVPR52734.2025.02245},
  bibsource    = {dblp computer science bibliography, https://dblp.org}
}

@misc{wang2025lvbenchextremelongvideo,
      title={LVBench: An Extreme Long Video Understanding Benchmark}, 
      author={Weihan Wang and Zehai He and Wenyi Hong and Yean Cheng and Xiaohan Zhang and Ji Qi and Xiaotao Gu and Shiyu Huang and Bin Xu and Yuxiao Dong and Ming Ding and Jie Tang},
      year={2025},
      eprint={2406.08035},
      archivePrefix={arXiv},
      primaryClass={cs.CV},
      url={https://arxiv.org/abs/2406.08035}, 
}

@misc{openai2024gpt4ocard,
      title={GPT-4o System Card}, 
      author={OpenAI and : and Aaron Hurst and Adam Lerer and Adam P. Goucher and Adam Perelman and Aditya Ramesh and Aidan Clark and AJ Ostrow and Akila Welihinda and Alan Hayes and Alec Radford and Aleksander Mądry and Alex Baker-Whitcomb and Alex Beutel and Alex Borzunov and Alex Carney and Alex Chow and Alex Kirillov and Alex Nichol and Alex Paino and Alex Renzin and Alex Tachard Passos and Alexander Kirillov and Alexi Christakis and Alexis Conneau and Ali Kamali and Allan Jabri and Allison Moyer and Allison Tam and Amadou Crookes and Amin Tootoochian and Amin Tootoonchian and Ananya Kumar and Andrea Vallone and Andrej Karpathy and Andrew Braunstein and Andrew Cann and Andrew Codispoti and Andrew Galu and Andrew Kondrich and Andrew Tulloch and Andrey Mishchenko and Angela Baek and Angela Jiang and Antoine Pelisse and Antonia Woodford and Anuj Gosalia and Arka Dhar and Ashley Pantuliano and Avi Nayak and Avital Oliver and Barret Zoph and Behrooz Ghorbani and Ben Leimberger and Ben Rossen and Ben Sokolowsky and Ben Wang and Benjamin Zweig and Beth Hoover and Blake Samic and Bob McGrew and Bobby Spero and Bogo Giertler and Bowen Cheng and Brad Lightcap and Brandon Walkin and Brendan Quinn and Brian Guarraci and Brian Hsu and Bright Kellogg and Brydon Eastman and Camillo Lugaresi and Carroll Wainwright and Cary Bassin and Cary Hudson and Casey Chu and Chad Nelson and Chak Li and Chan Jun Shern and Channing Conger and Charlotte Barette and Chelsea Voss and Chen Ding and Cheng Lu and Chong Zhang and Chris Beaumont and Chris Hallacy and Chris Koch and Christian Gibson and Christina Kim and Christine Choi and Christine McLeavey and Christopher Hesse and Claudia Fischer and Clemens Winter and Coley Czarnecki and Colin Jarvis and Colin Wei and Constantin Koumouzelis and Dane Sherburn and Daniel Kappler and Daniel Levin and Daniel Levy and David Carr and David Farhi and David Mely and David Robinson and David Sasaki and Denny Jin and Dev Valladares and Dimitris Tsipras and Doug Li and Duc Phong Nguyen and Duncan Findlay and Edede Oiwoh and Edmund Wong and Ehsan Asdar and Elizabeth Proehl and Elizabeth Yang and Eric Antonow and Eric Kramer and Eric Peterson and Eric Sigler and Eric Wallace and Eugene Brevdo and Evan Mays and Farzad Khorasani and Felipe Petroski Such and Filippo Raso and Francis Zhang and Fred von Lohmann and Freddie Sulit and Gabriel Goh and Gene Oden and Geoff Salmon and Giulio Starace and Greg Brockman and Hadi Salman and Haiming Bao and Haitang Hu and Hannah Wong and Haoyu Wang and Heather Schmidt and Heather Whitney and Heewoo Jun and Hendrik Kirchner and Henrique Ponde de Oliveira Pinto and Hongyu Ren and Huiwen Chang and Hyung Won Chung and Ian Kivlichan and Ian O'Connell and Ian O'Connell and Ian Osband and Ian Silber and Ian Sohl and Ibrahim Okuyucu and Ikai Lan and Ilya Kostrikov and Ilya Sutskever and Ingmar Kanitscheider and Ishaan Gulrajani and Jacob Coxon and Jacob Menick and Jakub Pachocki and James Aung and James Betker and James Crooks and James Lennon and Jamie Kiros and Jan Leike and Jane Park and Jason Kwon and Jason Phang and Jason Teplitz and Jason Wei and Jason Wolfe and Jay Chen and Jeff Harris and Jenia Varavva and Jessica Gan Lee and Jessica Shieh and Ji Lin and Jiahui Yu and Jiayi Weng and Jie Tang and Jieqi Yu and Joanne Jang and Joaquin Quinonero Candela and Joe Beutler and Joe Landers and Joel Parish and Johannes Heidecke and John Schulman and Jonathan Lachman and Jonathan McKay and Jonathan Uesato and Jonathan Ward and Jong Wook Kim and Joost Huizinga and Jordan Sitkin and Jos Kraaijeveld and Josh Gross and Josh Kaplan and Josh Snyder and Joshua Achiam and Joy Jiao and Joyce Lee and Juntang Zhuang and Justyn Harriman and Kai Fricke and Kai Hayashi and Karan Singhal and Katy Shi and Kavin Karthik and Kayla Wood and Kendra Rimbach and Kenny Hsu and Kenny Nguyen and Keren Gu-Lemberg and Kevin Button and Kevin Liu and Kiel Howe and Krithika Muthukumar and Kyle Luther and Lama Ahmad and Larry Kai and Lauren Itow and Lauren Workman and Leher Pathak and Leo Chen and Li Jing and Lia Guy and Liam Fedus and Liang Zhou and Lien Mamitsuka and Lilian Weng and Lindsay McCallum and Lindsey Held and Long Ouyang and Louis Feuvrier and Lu Zhang and Lukas Kondraciuk and Lukasz Kaiser and Luke Hewitt and Luke Metz and Lyric Doshi and Mada Aflak and Maddie Simens and Madelaine Boyd and Madeleine Thompson and Marat Dukhan and Mark Chen and Mark Gray and Mark Hudnall and Marvin Zhang and Marwan Aljubeh and Mateusz Litwin and Matthew Zeng and Max Johnson and Maya Shetty and Mayank Gupta and Meghan Shah and Mehmet Yatbaz and Meng Jia Yang and Mengchao Zhong and Mia Glaese and Mianna Chen and Michael Janner and Michael Lampe and Michael Petrov and Michael Wu and Michele Wang and Michelle Fradin and Michelle Pokrass and Miguel Castro and Miguel Oom Temudo de Castro and Mikhail Pavlov and Miles Brundage and Miles Wang and Minal Khan and Mira Murati and Mo Bavarian and Molly Lin and Murat Yesildal and Nacho Soto and Natalia Gimelshein and Natalie Cone and Natalie Staudacher and Natalie Summers and Natan LaFontaine and Neil Chowdhury and Nick Ryder and Nick Stathas and Nick Turley and Nik Tezak and Niko Felix and Nithanth Kudige and Nitish Keskar and Noah Deutsch and Noel Bundick and Nora Puckett and Ofir Nachum and Ola Okelola and Oleg Boiko and Oleg Murk and Oliver Jaffe and Olivia Watkins and Olivier Godement and Owen Campbell-Moore and Patrick Chao and Paul McMillan and Pavel Belov and Peng Su and Peter Bak and Peter Bakkum and Peter Deng and Peter Dolan and Peter Hoeschele and Peter Welinder and Phil Tillet and Philip Pronin and Philippe Tillet and Prafulla Dhariwal and Qiming Yuan and Rachel Dias and Rachel Lim and Rahul Arora and Rajan Troll and Randall Lin and Rapha Gontijo Lopes and Raul Puri and Reah Miyara and Reimar Leike and Renaud Gaubert and Reza Zamani and Ricky Wang and Rob Donnelly and Rob Honsby and Rocky Smith and Rohan Sahai and Rohit Ramchandani and Romain Huet and Rory Carmichael and Rowan Zellers and Roy Chen and Ruby Chen and Ruslan Nigmatullin and Ryan Cheu and Saachi Jain and Sam Altman and Sam Schoenholz and Sam Toizer and Samuel Miserendino and Sandhini Agarwal and Sara Culver and Scott Ethersmith and Scott Gray and Sean Grove and Sean Metzger and Shamez Hermani and Shantanu Jain and Shengjia Zhao and Sherwin Wu and Shino Jomoto and Shirong Wu and Shuaiqi and Xia and Sonia Phene and Spencer Papay and Srinivas Narayanan and Steve Coffey and Steve Lee and Stewart Hall and Suchir Balaji and Tal Broda and Tal Stramer and Tao Xu and Tarun Gogineni and Taya Christianson and Ted Sanders and Tejal Patwardhan and Thomas Cunninghman and Thomas Degry and Thomas Dimson and Thomas Raoux and Thomas Shadwell and Tianhao Zheng and Todd Underwood and Todor Markov and Toki Sherbakov and Tom Rubin and Tom Stasi and Tomer Kaftan and Tristan Heywood and Troy Peterson and Tyce Walters and Tyna Eloundou and Valerie Qi and Veit Moeller and Vinnie Monaco and Vishal Kuo and Vlad Fomenko and Wayne Chang and Weiyi Zheng and Wenda Zhou and Wesam Manassra and Will Sheu and Wojciech Zaremba and Yash Patil and Yilei Qian and Yongjik Kim and Youlong Cheng and Yu Zhang and Yuchen He and Yuchen Zhang and Yujia Jin and Yunxing Dai and Yury Malkov},
      year={2024},
      eprint={2410.21276},
      archivePrefix={arXiv},
      primaryClass={cs.CL},
      url={https://arxiv.org/abs/2410.21276}, 
}

@misc{geminiteam2024gemini15unlockingmultimodal,
      title={Gemini 1.5: Unlocking multimodal understanding across millions of tokens of context}, 
      author={Gemini Team and Petko Georgiev and Ving Ian Lei and Ryan Burnell and Libin Bai and Anmol Gulati and Garrett Tanzer and Damien Vincent and Zhufeng Pan and Shibo Wang and Soroosh Mariooryad and Yifan Ding and Xinyang Geng and Fred Alcober and Roy Frostig and Mark Omernick and Lexi Walker and Cosmin Paduraru and Christina Sorokin and Andrea Tacchetti and Colin Gaffney and Samira Daruki and Olcan Sercinoglu and Zach Gleicher and Juliette Love and Paul Voigtlaender and Rohan Jain and Gabriela Surita and Kareem Mohamed and Rory Blevins and Junwhan Ahn and Tao Zhu and Kornraphop Kawintiranon and Orhan Firat and Yiming Gu and Yujing Zhang and Matthew Rahtz and Manaal Faruqui and Natalie Clay and Justin Gilmer and JD Co-Reyes and Ivo Penchev and Rui Zhu and Nobuyuki Morioka and Kevin Hui and Krishna Haridasan and Victor Campos and Mahdis Mahdieh and Mandy Guo and Samer Hassan and Kevin Kilgour and Arpi Vezer and Heng-Tze Cheng and Raoul de Liedekerke and Siddharth Goyal and Paul Barham and DJ Strouse and Seb Noury and Jonas Adler and Mukund Sundararajan and Sharad Vikram and Dmitry Lepikhin and Michela Paganini and Xavier Garcia and Fan Yang and Dasha Valter and Maja Trebacz and Kiran Vodrahalli and Chulayuth Asawaroengchai and Roman Ring and Norbert Kalb and Livio Baldini Soares and Siddhartha Brahma and David Steiner and Tianhe Yu and Fabian Mentzer and Antoine He and Lucas Gonzalez and Bibo Xu and Raphael Lopez Kaufman and Laurent El Shafey and Junhyuk Oh and Tom Hennigan and George van den Driessche and Seth Odoom and Mario Lucic and Becca Roelofs and Sid Lall and Amit Marathe and Betty Chan and Santiago Ontanon and Luheng He and Denis Teplyashin and Jonathan Lai and Phil Crone and Bogdan Damoc and Lewis Ho and Sebastian Riedel and Karel Lenc and Chih-Kuan Yeh and Aakanksha Chowdhery and Yang Xu and Mehran Kazemi and Ehsan Amid and Anastasia Petrushkina and Kevin Swersky and Ali Khodaei and Gowoon Chen and Chris Larkin and Mario Pinto and Geng Yan and Adria Puigdomenech Badia and Piyush Patil and Steven Hansen and Dave Orr and Sebastien M. R. Arnold and Jordan Grimstad and Andrew Dai and Sholto Douglas and Rishika Sinha and Vikas Yadav and Xi Chen and Elena Gribovskaya and Jacob Austin and Jeffrey Zhao and Kaushal Patel and Paul Komarek and Sophia Austin and Sebastian Borgeaud and Linda Friso and Abhimanyu Goyal and Ben Caine and Kris Cao and Da-Woon Chung and Matthew Lamm and Gabe Barth-Maron and Thais Kagohara and Kate Olszewska and Mia Chen and Kaushik Shivakumar and Rishabh Agarwal and Harshal Godhia and Ravi Rajwar and Javier Snaider and Xerxes Dotiwalla and Yuan Liu and Aditya Barua and Victor Ungureanu and Yuan Zhang and Bat-Orgil Batsaikhan and Mateo Wirth and James Qin and Ivo Danihelka and Tulsee Doshi and Martin Chadwick and Jilin Chen and Sanil Jain and Quoc Le and Arjun Kar and Madhu Gurumurthy and Cheng Li and Ruoxin Sang and Fangyu Liu and Lampros Lamprou and Rich Munoz and Nathan Lintz and Harsh Mehta and Heidi Howard and Malcolm Reynolds and Lora Aroyo and Quan Wang and Lorenzo Blanco and Albin Cassirer and Jordan Griffith and Dipanjan Das and Stephan Lee and Jakub Sygnowski and Zach Fisher and James Besley and Richard Powell and Zafarali Ahmed and Dominik Paulus and David Reitter and Zalan Borsos and Rishabh Joshi and Aedan Pope and Steven Hand and Vittorio Selo and Vihan Jain and Nikhil Sethi and Megha Goel and Takaki Makino and Rhys May and Zhen Yang and Johan Schalkwyk and Christina Butterfield and Anja Hauth and Alex Goldin and Will Hawkins and Evan Senter and Sergey Brin and Oliver Woodman and Marvin Ritter and Eric Noland and Minh Giang and Vijay Bolina and Lisa Lee and Tim Blyth and Ian Mackinnon and Machel Reid and Obaid Sarvana and David Silver and Alexander Chen and Lily Wang and Loren Maggiore and Oscar Chang and Nithya Attaluri and Gregory Thornton and Chung-Cheng Chiu and Oskar Bunyan and Nir Levine and Timothy Chung and Evgenii Eltyshev and Xiance Si and Timothy Lillicrap and Demetra Brady and Vaibhav Aggarwal and Boxi Wu and Yuanzhong Xu and Ross McIlroy and Kartikeya Badola and Paramjit Sandhu and Erica Moreira and Wojciech Stokowiec and Ross Hemsley and Dong Li and Alex Tudor and Pranav Shyam and Elahe Rahimtoroghi and Salem Haykal and Pablo Sprechmann and Xiang Zhou and Diana Mincu and Yujia Li and Ravi Addanki and Kalpesh Krishna and Xiao Wu and Alexandre Frechette and Matan Eyal and Allan Dafoe and Dave Lacey and Jay Whang and Thi Avrahami and Ye Zhang and Emanuel Taropa and Hanzhao Lin and Daniel Toyama and Eliza Rutherford and Motoki Sano and HyunJeong Choe and Alex Tomala and Chalence Safranek-Shrader and Nora Kassner and Mantas Pajarskas and Matt Harvey and Sean Sechrist and Meire Fortunato and Christina Lyu and Gamaleldin Elsayed and Chenkai Kuang and James Lottes and Eric Chu and Chao Jia and Chih-Wei Chen and Peter Humphreys and Kate Baumli and Connie Tao and Rajkumar Samuel and Cicero Nogueira dos Santos and Anders Andreassen and Nemanja Rakićević and Dominik Grewe and Aviral Kumar and Stephanie Winkler and Jonathan Caton and Andrew Brock and Sid Dalmia and Hannah Sheahan and Iain Barr and Yingjie Miao and Paul Natsev and Jacob Devlin and Feryal Behbahani and Flavien Prost and Yanhua Sun and Artiom Myaskovsky and Thanumalayan Sankaranarayana Pillai and Dan Hurt and Angeliki Lazaridou and Xi Xiong and Ce Zheng and Fabio Pardo and Xiaowei Li and Dan Horgan and Joe Stanton and Moran Ambar and Fei Xia and Alejandro Lince and Mingqiu Wang and Basil Mustafa and Albert Webson and Hyo Lee and Rohan Anil and Martin Wicke and Timothy Dozat and Abhishek Sinha and Enrique Piqueras and Elahe Dabir and Shyam Upadhyay and Anudhyan Boral and Lisa Anne Hendricks and Corey Fry and Josip Djolonga and Yi Su and Jake Walker and Jane Labanowski and Ronny Huang and Vedant Misra and Jeremy Chen and RJ Skerry-Ryan and Avi Singh and Shruti Rijhwani and Dian Yu and Alex Castro-Ros and Beer Changpinyo and Romina Datta and Sumit Bagri and Arnar Mar Hrafnkelsson and Marcello Maggioni and Daniel Zheng and Yury Sulsky and Shaobo Hou and Tom Le Paine and Antoine Yang and Jason Riesa and Dominika Rogozinska and Dror Marcus and Dalia El Badawy and Qiao Zhang and Luyu Wang and Helen Miller and Jeremy Greer and Lars Lowe Sjos and Azade Nova and Heiga Zen and Rahma Chaabouni and Mihaela Rosca and Jiepu Jiang and Charlie Chen and Ruibo Liu and Tara Sainath and Maxim Krikun and Alex Polozov and Jean-Baptiste Lespiau and Josh Newlan and Zeyncep Cankara and Soo Kwak and Yunhan Xu and Phil Chen and Andy Coenen and Clemens Meyer and Katerina Tsihlas and Ada Ma and Juraj Gottweis and Jinwei Xing and Chenjie Gu and Jin Miao and Christian Frank and Zeynep Cankara and Sanjay Ganapathy and Ishita Dasgupta and Steph Hughes-Fitt and Heng Chen and David Reid and Keran Rong and Hongmin Fan and Joost van Amersfoort and Vincent Zhuang and Aaron Cohen and Shixiang Shane Gu and Anhad Mohananey and Anastasija Ilic and Taylor Tobin and John Wieting and Anna Bortsova and Phoebe Thacker and Emma Wang and Emily Caveness and Justin Chiu and Eren Sezener and Alex Kaskasoli and Steven Baker and Katie Millican and Mohamed Elhawaty and Kostas Aisopos and Carl Lebsack and Nathan Byrd and Hanjun Dai and Wenhao Jia and Matthew Wiethoff and Elnaz Davoodi and Albert Weston and Lakshman Yagati and Arun Ahuja and Isabel Gao and Golan Pundak and Susan Zhang and Michael Azzam and Khe Chai Sim and Sergi Caelles and James Keeling and Abhanshu Sharma and Andy Swing and YaGuang Li and Chenxi Liu and Carrie Grimes Bostock and Yamini Bansal and Zachary Nado and Ankesh Anand and Josh Lipschultz and Abhijit Karmarkar and Lev Proleev and Abe Ittycheriah and Soheil Hassas Yeganeh and George Polovets and Aleksandra Faust and Jiao Sun and Alban Rrustemi and Pen Li and Rakesh Shivanna and Jeremiah Liu and Chris Welty and Federico Lebron and Anirudh Baddepudi and Sebastian Krause and Emilio Parisotto and Radu Soricut and Zheng Xu and Dawn Bloxwich and Melvin Johnson and Behnam Neyshabur and Justin Mao-Jones and Renshen Wang and Vinay Ramasesh and Zaheer Abbas and Arthur Guez and Constant Segal and Duc Dung Nguyen and James Svensson and Le Hou and Sarah York and Kieran Milan and Sophie Bridgers and Wiktor Gworek and Marco Tagliasacchi and James Lee-Thorp and Michael Chang and Alexey Guseynov and Ale Jakse Hartman and Michael Kwong and Ruizhe Zhao and Sheleem Kashem and Elizabeth Cole and Antoine Miech and Richard Tanburn and Mary Phuong and Filip Pavetic and Sebastien Cevey and Ramona Comanescu and Richard Ives and Sherry Yang and Cosmo Du and Bo Li and Zizhao Zhang and Mariko Iinuma and Clara Huiyi Hu and Aurko Roy and Shaan Bijwadia and Zhenkai Zhu and Danilo Martins and Rachel Saputro and Anita Gergely and Steven Zheng and Dawei Jia and Ioannis Antonoglou and Adam Sadovsky and Shane Gu and Yingying Bi and Alek Andreev and Sina Samangooei and Mina Khan and Tomas Kocisky and Angelos Filos and Chintu Kumar and Colton Bishop and Adams Yu and Sarah Hodkinson and Sid Mittal and Premal Shah and Alexandre Moufarek and Yong Cheng and Adam Bloniarz and Jaehoon Lee and Pedram Pejman and Paul Michel and Stephen Spencer and Vladimir Feinberg and Xuehan Xiong and Nikolay Savinov and Charlotte Smith and Siamak Shakeri and Dustin Tran and Mary Chesus and Bernd Bohnet and George Tucker and Tamara von Glehn and Carrie Muir and Yiran Mao and Hideto Kazawa and Ambrose Slone and Kedar Soparkar and Disha Shrivastava and James Cobon-Kerr and Michael Sharman and Jay Pavagadhi and Carlos Araya and Karolis Misiunas and Nimesh Ghelani and Michael Laskin and David Barker and Qiujia Li and Anton Briukhov and Neil Houlsby and Mia Glaese and Balaji Lakshminarayanan and Nathan Schucher and Yunhao Tang and Eli Collins and Hyeontaek Lim and Fangxiaoyu Feng and Adria Recasens and Guangda Lai and Alberto Magni and Nicola De Cao and Aditya Siddhant and Zoe Ashwood and Jordi Orbay and Mostafa Dehghani and Jenny Brennan and Yifan He and Kelvin Xu and Yang Gao and Carl Saroufim and James Molloy and Xinyi Wu and Seb Arnold and Solomon Chang and Julian Schrittwieser and Elena Buchatskaya and Soroush Radpour and Martin Polacek and Skye Giordano and Ankur Bapna and Simon Tokumine and Vincent Hellendoorn and Thibault Sottiaux and Sarah Cogan and Aliaksei Severyn and Mohammad Saleh and Shantanu Thakoor and Laurent Shefey and Siyuan Qiao and Meenu Gaba and Shuo-yiin Chang and Craig Swanson and Biao Zhang and Benjamin Lee and Paul Kishan Rubenstein and Gan Song and Tom Kwiatkowski and Anna Koop and Ajay Kannan and David Kao and Parker Schuh and Axel Stjerngren and Golnaz Ghiasi and Gena Gibson and Luke Vilnis and Ye Yuan and Felipe Tiengo Ferreira and Aishwarya Kamath and Ted Klimenko and Ken Franko and Kefan Xiao and Indro Bhattacharya and Miteyan Patel and Rui Wang and Alex Morris and Robin Strudel and Vivek Sharma and Peter Choy and Sayed Hadi Hashemi and Jessica Landon and Mara Finkelstein and Priya Jhakra and Justin Frye and Megan Barnes and Matthew Mauger and Dennis Daun and Khuslen Baatarsukh and Matthew Tung and Wael Farhan and Henryk Michalewski and Fabio Viola and Felix de Chaumont Quitry and Charline Le Lan and Tom Hudson and Qingze Wang and Felix Fischer and Ivy Zheng and Elspeth White and Anca Dragan and Jean-baptiste Alayrac and Eric Ni and Alexander Pritzel and Adam Iwanicki and Michael Isard and Anna Bulanova and Lukas Zilka and Ethan Dyer and Devendra Sachan and Srivatsan Srinivasan and Hannah Muckenhirn and Honglong Cai and Amol Mandhane and Mukarram Tariq and Jack W. Rae and Gary Wang and Kareem Ayoub and Nicholas FitzGerald and Yao Zhao and Woohyun Han and Chris Alberti and Dan Garrette and Kashyap Krishnakumar and Mai Gimenez and Anselm Levskaya and Daniel Sohn and Josip Matak and Inaki Iturrate and Michael B. Chang and Jackie Xiang and Yuan Cao and Nishant Ranka and Geoff Brown and Adrian Hutter and Vahab Mirrokni and Nanxin Chen and Kaisheng Yao and Zoltan Egyed and Francois Galilee and Tyler Liechty and Praveen Kallakuri and Evan Palmer and Sanjay Ghemawat and Jasmine Liu and David Tao and Chloe Thornton and Tim Green and Mimi Jasarevic and Sharon Lin and Victor Cotruta and Yi-Xuan Tan and Noah Fiedel and Hongkun Yu and Ed Chi and Alexander Neitz and Jens Heitkaemper and Anu Sinha and Denny Zhou and Yi Sun and Charbel Kaed and Brice Hulse and Swaroop Mishra and Maria Georgaki and Sneha Kudugunta and Clement Farabet and Izhak Shafran and Daniel Vlasic and Anton Tsitsulin and Rajagopal Ananthanarayanan and Alen Carin and Guolong Su and Pei Sun and Shashank V and Gabriel Carvajal and Josef Broder and Iulia Comsa and Alena Repina and William Wong and Warren Weilun Chen and Peter Hawkins and Egor Filonov and Lucia Loher and Christoph Hirnschall and Weiyi Wang and Jingchen Ye and Andrea Burns and Hardie Cate and Diana Gage Wright and Federico Piccinini and Lei Zhang and Chu-Cheng Lin and Ionel Gog and Yana Kulizhskaya and Ashwin Sreevatsa and Shuang Song and Luis C. Cobo and Anand Iyer and Chetan Tekur and Guillermo Garrido and Zhuyun Xiao and Rupert Kemp and Huaixiu Steven Zheng and Hui Li and Ananth Agarwal and Christel Ngani and Kati Goshvadi and Rebeca Santamaria-Fernandez and Wojciech Fica and Xinyun Chen and Chris Gorgolewski and Sean Sun and Roopal Garg and Xinyu Ye and S. M. Ali Eslami and Nan Hua and Jon Simon and Pratik Joshi and Yelin Kim and Ian Tenney and Sahitya Potluri and Lam Nguyen Thiet and Quan Yuan and Florian Luisier and Alexandra Chronopoulou and Salvatore Scellato and Praveen Srinivasan and Minmin Chen and Vinod Koverkathu and Valentin Dalibard and Yaming Xu and Brennan Saeta and Keith Anderson and Thibault Sellam and Nick Fernando and Fantine Huot and Junehyuk Jung and Mani Varadarajan and Michael Quinn and Amit Raul and Maigo Le and Ruslan Habalov and Jon Clark and Komal Jalan and Kalesha Bullard and Achintya Singhal and Thang Luong and Boyu Wang and Sujeevan Rajayogam and Julian Eisenschlos and Johnson Jia and Daniel Finchelstein and Alex Yakubovich and Daniel Balle and Michael Fink and Sameer Agarwal and Jing Li and Dj Dvijotham and Shalini Pal and Kai Kang and Jaclyn Konzelmann and Jennifer Beattie and Olivier Dousse and Diane Wu and Remi Crocker and Chen Elkind and Siddhartha Reddy Jonnalagadda and Jong Lee and Dan Holtmann-Rice and Krystal Kallarackal and Rosanne Liu and Denis Vnukov and Neera Vats and Luca Invernizzi and Mohsen Jafari and Huanjie Zhou and Lilly Taylor and Jennifer Prendki and Marcus Wu and Tom Eccles and Tianqi Liu and Kavya Kopparapu and Francoise Beaufays and Christof Angermueller and Andreea Marzoca and Shourya Sarcar and Hilal Dib and Jeff Stanway and Frank Perbet and Nejc Trdin and Rachel Sterneck and Andrey Khorlin and Dinghua Li and Xihui Wu and Sonam Goenka and David Madras and Sasha Goldshtein and Willi Gierke and Tong Zhou and Yaxin Liu and Yannie Liang and Anais White and Yunjie Li and Shreya Singh and Sanaz Bahargam and Mark Epstein and Sujoy Basu and Li Lao and Adnan Ozturel and Carl Crous and Alex Zhai and Han Lu and Zora Tung and Neeraj Gaur and Alanna Walton and Lucas Dixon and Ming Zhang and Amir Globerson and Grant Uy and Andrew Bolt and Olivia Wiles and Milad Nasr and Ilia Shumailov and Marco Selvi and Francesco Piccinno and Ricardo Aguilar and Sara McCarthy and Misha Khalman and Mrinal Shukla and Vlado Galic and John Carpenter and Kevin Villela and Haibin Zhang and Harry Richardson and James Martens and Matko Bosnjak and Shreyas Rammohan Belle and Jeff Seibert and Mahmoud Alnahlawi and Brian McWilliams and Sankalp Singh and Annie Louis and Wen Ding and Dan Popovici and Lenin Simicich and Laura Knight and Pulkit Mehta and Nishesh Gupta and Chongyang Shi and Saaber Fatehi and Jovana Mitrovic and Alex Grills and Joseph Pagadora and Tsendsuren Munkhdalai and Dessie Petrova and Danielle Eisenbud and Zhishuai Zhang and Damion Yates and Bhavishya Mittal and Nilesh Tripuraneni and Yannis Assael and Thomas Brovelli and Prateek Jain and Mihajlo Velimirovic and Canfer Akbulut and Jiaqi Mu and Wolfgang Macherey and Ravin Kumar and Jun Xu and Haroon Qureshi and Gheorghe Comanici and Jeremy Wiesner and Zhitao Gong and Anton Ruddock and Matthias Bauer and Nick Felt and Anirudh GP and Anurag Arnab and Dustin Zelle and Jonas Rothfuss and Bill Rosgen and Ashish Shenoy and Bryan Seybold and Xinjian Li and Jayaram Mudigonda and Goker Erdogan and Jiawei Xia and Jiri Simsa and Andrea Michi and Yi Yao and Christopher Yew and Steven Kan and Isaac Caswell and Carey Radebaugh and Andre Elisseeff and Pedro Valenzuela and Kay McKinney and Kim Paterson and Albert Cui and Eri Latorre-Chimoto and Solomon Kim and William Zeng and Ken Durden and Priya Ponnapalli and Tiberiu Sosea and Christopher A. Choquette-Choo and James Manyika and Brona Robenek and Harsha Vashisht and Sebastien Pereira and Hoi Lam and Marko Velic and Denese Owusu-Afriyie and Katherine Lee and Tolga Bolukbasi and Alicia Parrish and Shawn Lu and Jane Park and Balaji Venkatraman and Alice Talbert and Lambert Rosique and Yuchung Cheng and Andrei Sozanschi and Adam Paszke and Praveen Kumar and Jessica Austin and Lu Li and Khalid Salama and Bartek Perz and Wooyeol Kim and Nandita Dukkipati and Anthony Baryshnikov and Christos Kaplanis and XiangHai Sheng and Yuri Chervonyi and Caglar Unlu and Diego de Las Casas and Harry Askham and Kathryn Tunyasuvunakool and Felix Gimeno and Siim Poder and Chester Kwak and Matt Miecnikowski and Vahab Mirrokni and Alek Dimitriev and Aaron Parisi and Dangyi Liu and Tomy Tsai and Toby Shevlane and Christina Kouridi and Drew Garmon and Adrian Goedeckemeyer and Adam R. Brown and Anitha Vijayakumar and Ali Elqursh and Sadegh Jazayeri and Jin Huang and Sara Mc Carthy and Jay Hoover and Lucy Kim and Sandeep Kumar and Wei Chen and Courtney Biles and Garrett Bingham and Evan Rosen and Lisa Wang and Qijun Tan and David Engel and Francesco Pongetti and Dario de Cesare and Dongseong Hwang and Lily Yu and Jennifer Pullman and Srini Narayanan and Kyle Levin and Siddharth Gopal and Megan Li and Asaf Aharoni and Trieu Trinh and Jessica Lo and Norman Casagrande and Roopali Vij and Loic Matthey and Bramandia Ramadhana and Austin Matthews and CJ Carey and Matthew Johnson and Kremena Goranova and Rohin Shah and Shereen Ashraf and Kingshuk Dasgupta and Rasmus Larsen and Yicheng Wang and Manish Reddy Vuyyuru and Chong Jiang and Joana Ijazi and Kazuki Osawa and Celine Smith and Ramya Sree Boppana and Taylan Bilal and Yuma Koizumi and Ying Xu and Yasemin Altun and Nir Shabat and Ben Bariach and Alex Korchemniy and Kiam Choo and Olaf Ronneberger and Chimezie Iwuanyanwu and Shubin Zhao and David Soergel and Cho-Jui Hsieh and Irene Cai and Shariq Iqbal and Martin Sundermeyer and Zhe Chen and Elie Bursztein and Chaitanya Malaviya and Fadi Biadsy and Prakash Shroff and Inderjit Dhillon and Tejasi Latkar and Chris Dyer and Hannah Forbes and Massimo Nicosia and Vitaly Nikolaev and Somer Greene and Marin Georgiev and Pidong Wang and Nina Martin and Hanie Sedghi and John Zhang and Praseem Banzal and Doug Fritz and Vikram Rao and Xuezhi Wang and Jiageng Zhang and Viorica Patraucean and Dayou Du and Igor Mordatch and Ivan Jurin and Lewis Liu and Ayush Dubey and Abhi Mohan and Janek Nowakowski and Vlad-Doru Ion and Nan Wei and Reiko Tojo and Maria Abi Raad and Drew A. Hudson and Vaishakh Keshava and Shubham Agrawal and Kevin Ramirez and Zhichun Wu and Hoang Nguyen and Ji Liu and Madhavi Sewak and Bryce Petrini and DongHyun Choi and Ivan Philips and Ziyue Wang and Ioana Bica and Ankush Garg and Jarek Wilkiewicz and Priyanka Agrawal and Xiaowei Li and Danhao Guo and Emily Xue and Naseer Shaik and Andrew Leach and Sadh MNM Khan and Julia Wiesinger and Sammy Jerome and Abhishek Chakladar and Alek Wenjiao Wang and Tina Ornduff and Folake Abu and Alireza Ghaffarkhah and Marcus Wainwright and Mario Cortes and Frederick Liu and Joshua Maynez and Andreas Terzis and Pouya Samangouei and Riham Mansour and Tomasz Kępa and François-Xavier Aubet and Anton Algymr and Dan Banica and Agoston Weisz and Andras Orban and Alexandre Senges and Ewa Andrejczuk and Mark Geller and Niccolo Dal Santo and Valentin Anklin and Majd Al Merey and Martin Baeuml and Trevor Strohman and Junwen Bai and Slav Petrov and Yonghui Wu and Demis Hassabis and Koray Kavukcuoglu and Jeff Dean and Oriol Vinyals},
      year={2024},
      eprint={2403.05530},
      archivePrefix={arXiv},
      primaryClass={cs.CL},
      url={https://arxiv.org/abs/2403.05530}, 
}

@misc{dao2022flashattentionfastmemoryefficientexact,
      title={FlashAttention: Fast and Memory-Efficient Exact Attention with IO-Awareness}, 
      author={Tri Dao and Daniel Y. Fu and Stefano Ermon and Atri Rudra and Christopher Ré},
      year={2022},
      eprint={2205.14135},
      archivePrefix={arXiv},
      primaryClass={cs.LG},
      url={https://arxiv.org/abs/2205.14135}, 
}

@misc{rabe2022selfattentiondoesneedon2,
      title={Self-attention Does Not Need $O(n^2)$ Memory}, 
      author={Markus N. Rabe and Charles Staats},
      year={2022},
      eprint={2112.05682},
      archivePrefix={arXiv},
      primaryClass={cs.LG},
      url={https://arxiv.org/abs/2112.05682}, 
}

@inproceedings{DBLP:conf/cvpr/0001DD0ZCL025,
  author       = {Rui Qian and
                  Shuangrui Ding and
                  Xiaoyi Dong and
                  Pan Zhang and
                  Yuhang Zang and
                  Yuhang Cao and
                  Dahua Lin and
                  Jiaqi Wang},
  title        = {Dispider: Enabling Video LLMs with Active Real-Time Interaction via
                  Disentangled Perception, Decision, and Reaction},
  booktitle    = {{IEEE/CVF} Conference on Computer Vision and Pattern Recognition,
                  {CVPR} 2025, Nashville, TN, USA, June 11-15, 2025},
  pages        = {24045--24055},
  publisher    = {Computer Vision Foundation / {IEEE}},
  year         = {2025},
  url          = {https://openaccess.thecvf.com/content/CVPR2025/html/Qian\_Dispider\_Enabling\_Video\_LLMs\_with\_Active\_Real-Time\_Interaction\_via\_Disentangled\_CVPR\_2025\_paper.html},
  doi          = {10.1109/CVPR52734.2025.02239},
  bibsource    = {dblp computer science bibliography, https://dblp.org}
}

@article{huang2024online,
  title={Online video understanding: A comprehensive benchmark and memory-augmented method},
  author={Huang, Zhenpeng and Li, Xinhao and Li, Jiaqi and Wang, Jing and Zeng, Xiangyu and Liang, Cheng and Wu, Tao and Chen, Xi and Li, Liang and Wang, Limin},
  journal={arXiv e-prints},
  pages={arXiv--2501},
  year={2024}
}

\clearpage
\appendix

\renewcommand{\thetable}{A\arabic{table}}
\renewcommand{\thefigure}{A\arabic{figure}}
\setcounter{table}{0}
\setcounter{figure}{0}

\section{Future Work}
Future work will explore adaptive segmentation that selects segment boundaries online based on scene changes and question demands, reducing redundant memory updates while preserving evidence coverage. We also plan to incorporate audio cues and speech transcripts to support richer streaming understanding in real-world settings. Another direction is improving robustness on very long streams through better uncertainty estimation, memory verification, and training with harder distractors and domain shifts. Finally, we will develop more comprehensive evaluation protocols that jointly measure accuracy, latency, and resource usage in multi-turn interaction.

\section{Implementation Details}

All experiments are conducted on 8$\times$ NVIDIA RTX A6000 GPUs (48GB each). We fine-tune Qwen3-VL-2/4/8B-Instruct using full-parameter supervised fine-tuning (SFT). Table~\ref{tab:impl_details} summarizes the main training settings.

\begin{table}[h]
\centering
\caption{Implementation details and training hyperparameters.}
\label{tab:impl_details}
\begin{tabular}{ll}
\toprule
\textbf{Item} & \textbf{Setting} \\
\midrule
GPUs & 8$\times$ NVIDIA RTX A6000 (48GB) \\
Model & Qwen3-VL-2/4/8B-Instruct \\
Training method & Full-parameter SFT \\
Precision & bf16 \\
Global batch size & 128 \\
Optimizer & AdamW \\
Peak learning rate & $1.0\times10^{-5}$ \\
Learning rate schedule & Cosine decay \\
Warmup ratio & 3\% \\
Weight decay & 0.1 \\
Stage 1 strategy & DeepSpeed ZeRO-3 \\
Gradient checkpointing & Enabled \\
Communication overlap & Enabled \\
\bottomrule
\end{tabular}
\end{table}

\section{Theoretical Latency Derivation}
\label{app:latency}

\begin{figure}[h]
  \centering
  \includegraphics[width=0.72\linewidth]{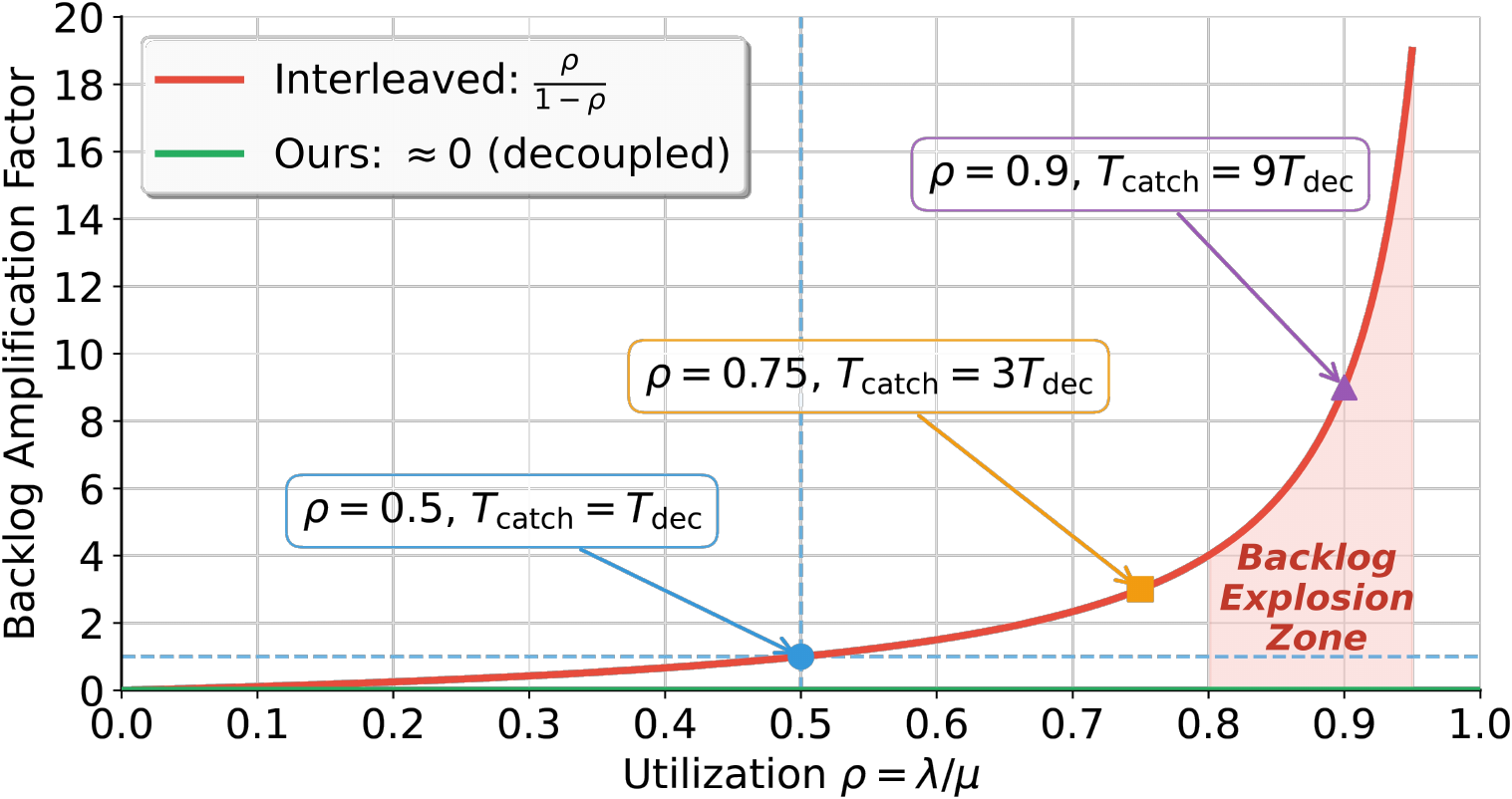}
  \caption{\textbf{Decoder-induced ingestion backlog under interleaved streaming.}
    As utilization $\rho$ increases, interleaved decoding pauses can amplify the catch-up delay and enter a backlog explosion regime, while our decoupled design substantially reduces decoder-induced backlog growth.}
  \label{fig:latency}
\end{figure}

We derive a simple queueing-style model to quantify decoder-induced ingestion backlog in interleaved streaming, as illustrated in Fig.~\ref{fig:latency}.

\noindent\textbf{Stream arrival and processing rates.}
Assume that video segments arrive in real time at a rate of $\lambda$ segments per second. When the model is watching, meaning that it is ingesting and prefilling, it processes segments at a rate of $\mu$ segments per second. We define the utilization by $\rho \triangleq \lambda/\mu$.

\noindent\textbf{Interleaved decoding as server downtime.}
In the interleaved baseline, generation is non-preemptive: during a decoding period of duration $T_{\mathrm{dec}}$, the system does not ingest new segments. While decoding, the stream continues to arrive, creating a backlog of
\begin{equation}
B \;=\; \lambda\, T_{\mathrm{dec}} .
\label{eq:backlog_size}
\end{equation}
$B$ represents backlog, which refers to the number of video segments that are accumulated and yet to be processed. After decoding, the system resumes watching at rate $\mu$ while new segments still arrive at rate $\lambda$. Thus the backlog drains at a net rate $(\mu-\lambda)$ (assuming $\mu>\lambda$), and the \textbf{catch-up time} is
\begin{equation}
T_{\mathrm{catch}}
\;=\;
\frac{B}{\mu-\lambda}
\;=\;
\frac{\lambda}{\mu-\lambda}\, T_{\mathrm{dec}}
\;=\;
\frac{\rho}{1-\rho}\, T_{\mathrm{dec}} .
\label{eq:catchup_time}
\end{equation}
This yields an amplification effect: each second spent decoding induces an additional $\rho/(1-\rho)$ second of future delay before the system fully catches up, which diverges as $\rho \rightarrow 1$.

\noindent\textbf{Think While Watching weakens backlog coupling.}
Our inference design decouples ingestion from decoding via dual KV caching, so decoding no longer forces a full stop of stream ingestion as in interleaved streaming. As a result, the decoder-induced backlog is greatly reduced. In an ideal fully overlapped implementation, the additional ingestion downtime during decoding approaches zero, yielding
\begin{equation}
B_{\mathrm{ours}} \approx 0 \quad\Rightarrow\quad T_{\mathrm{catch}} \approx 0 .
\end{equation}
In practice, however, a residual backlog may still arise from system overheads such as scheduling, synchronization, cache maintenance, and time overlapping. Therefore, the main benefit of our design is not to guarantee zero backlog but to substantially reduce the coupling between decoding and future stream lag.

\noindent\textbf{A quality real-time constraint in interleaving.}
Let $c_{\mathrm{tok}}$ be the average decoding time per output token and $L$ be the number of generated tokens per step. Interleaving spends $T_{\mathrm{dec}} = c_{\mathrm{tok}}L$ during which ingestion is paused, inducing
$T_{\mathrm{catch}}=\frac{\rho}{1-\rho}c_{\mathrm{tok}}L$ by Eq.~\eqref{eq:catchup_time}. Therefore, increasing $L$ to improve quality directly increases future stream lag. Our method weakens this coupling by allowing ingestion to proceed while decoding.

\section{Additional Baseline Details}
\label{app:baselines}

\noindent\textbf{Flash-VStream-7B\cite{DBLP:journals/corr/abs-2506-23825}.}
Flash-VStream is an efficient video language model for long video streams. It introduces a flash memory module composed of a low-capacity context memory for aggregating long-range temporal information and a high-capacity augmentation memory for retrieving detailed spatial evidence, enabling real-time responses to user queries over extremely long videos.

\noindent\textbf{VideoLLM-online-8B\cite{DBLP:conf/cvpr/ChenLWLSGLGMS24}.}
VideoLLM-online proposes the LIVE framework, short for learning in video streams, to enable temporally aligned, long-context, and real-time conversation over continuous video streams. Specifically, LIVE introduces a streaming training objective, a data generation scheme that converts offline temporal annotations into streaming dialogue data, and an optimized inference pipeline with a continuous KV cache as well as parallelized visual encoding and language decoding for efficient online responses.

\noindent\textbf{Dispider-7B\cite{DBLP:conf/cvpr/0001DD0ZCL025}.}
Dispider targets active real-time interaction by explicitly disentangling perception, decision, and reaction. It uses a lightweight proactive streaming video processing module to continuously monitor the stream and identify suitable moments for interaction, while an asynchronous interaction module generates detailed responses without blocking continued observation.

\noindent\textbf{StreamForest-7B\cite{DBLP:journals/corr/abs-2509-24871}.}
StreamForest is designed for efficient online video understanding with persistent event memory. Its core Persistent Event Memory Forest organizes historical frames into event-level tree structures for long-term retention under limited computational resources, while a Fine-grained Spatiotemporal Window preserves detailed short-term perception.

\noindent\textbf{StreamAgent-7B\cite{DBLP:journals/corr/abs-2508-01875}.}
StreamAgent studies anticipatory agents for streaming video understanding. Instead of reacting only to current observations, it integrates question semantics and historical observations to anticipate future task-relevant temporal intervals and spatial regions, and combines this strategy with a streaming KV cache memory for selective recall, enabling proactive and goal-directed responses in evolving video streams.

\section{Dataset and Benchmark Details}
\label{app:data}

\subsection{Benchmark Details}
\label{app:benchmark-details}

\noindent\textbf{StreamingBench~\cite{lin2024streamingbenchassessinggapmllms}.}
StreamingBench is a benchmark tailored for streaming video understanding, containing 18 tasks over 900 videos and 4,500 human-curated QA pairs, where each question is associated with a specific timestamp in the video stream. The benchmark covers three major aspects of streaming understanding: real-time visual understanding, omnisource understanding, and contextual understanding. In our main tables, we further summarize the reported results into four subset-level metrics: \textbf{Realtime}, \textbf{OmniSource}, \textbf{SQA}, and \textbf{Proactive}. For evaluation, we follow the adaptive frame extraction protocol reported in StreamingBench: videos shorter than 5 minutes are sampled at 1 fps, videos between 5 and 10 minutes at 0.5 fps, and videos longer than 10 minutes at 0.2 fps.

\noindent\textbf{OVO-Bench~\cite{DBLP:conf/cvpr/NiuLMGZHDDD0ZZC25}.}
OVO-Bench is designed to evaluate online video understanding with explicit temporal awareness. It organizes evaluation into three subsets: \textbf{Backward}, which requires tracing back to past events; \textbf{Realtime}, which focuses on understanding what is happening at the current timestamp; and \textbf{Forward}, which evaluates whether the model can defer its response until sufficient future evidence becomes available. In our experiments, we follow the common OVO-Bench comparison setting for offline video LLMs and cap the visual input at no more than 64 frames per query.

\noindent\textbf{Video-MME~\cite{DBLP:conf/cvpr/FuDLLRZWZSZCLLZ25}.}
Video-MME is a comprehensive offline video benchmark with 900 videos and 2,700 multiple-choice QA pairs, spanning 6 primary visual domains and 30 subfields, and reporting results on \textbf{Short}, \textbf{Medium}, and \textbf{Long} duration subsets. To adapt Video-MME to our streaming-style evaluation, we aggregate all QA pairs sharing the same video ID into a single example and convert each video into an ordered stream of temporal segments. The model receives these segments sequentially, and the associated questions are issued only after the full video stream has been observed. This preserves the original benchmark content while turning Video-MME into a suffix-query streaming protocol.

\noindent\textbf{LV-Bench~\cite{wang2025lvbenchextremelongvideo}.}
LV-Bench is an extreme long-video benchmark for evaluating long-range video understanding. It measures six core capabilities, namely \textbf{ER} (Entity Recognition), \textbf{EU} (Event Understanding), \textbf{KIR} (Key Information Retrieval), \textbf{TG} (Temporal Grounding), \textbf{Rea} (Reasoning), and \textbf{Sum} (Summarization). In our adaptation, we first aggregate samples by video ID, and then use the end time of the official annotated time span as the segment boundary to construct a streaming input sequence. The model is queried when the stream reaches the corresponding boundary, making the evaluation compatible with our streaming inference pipeline while preserving the original supervision structure.

\begin{figure*}[t]
  \centering
  \begin{minipage}[t]{0.32\textwidth}
    \centering
    \includegraphics[width=\linewidth]{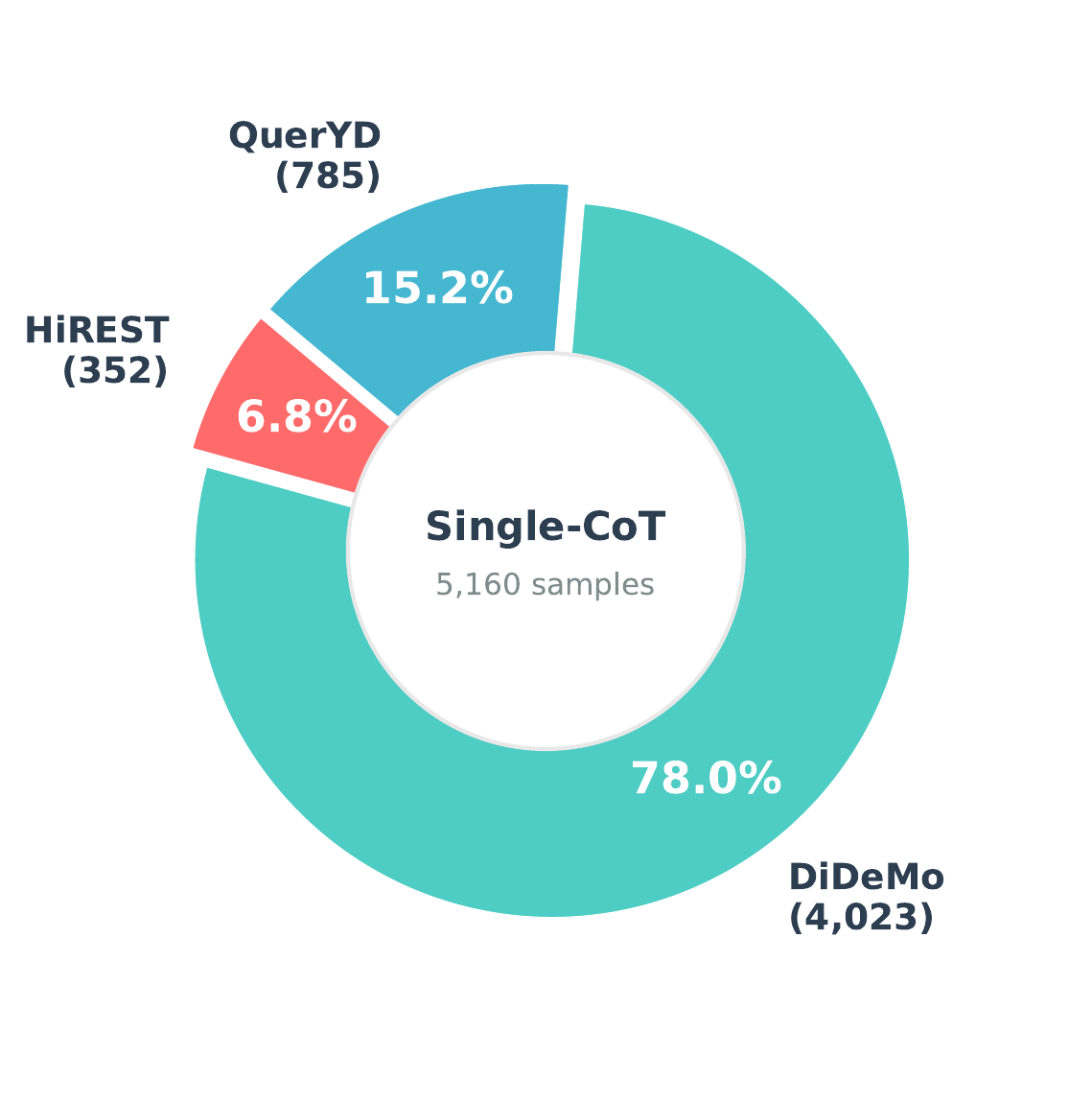}
    \par\small (a) Single-round CoT.
  \end{minipage}\hfill
  \begin{minipage}[t]{0.32\textwidth}
    \centering
    \includegraphics[width=\linewidth]{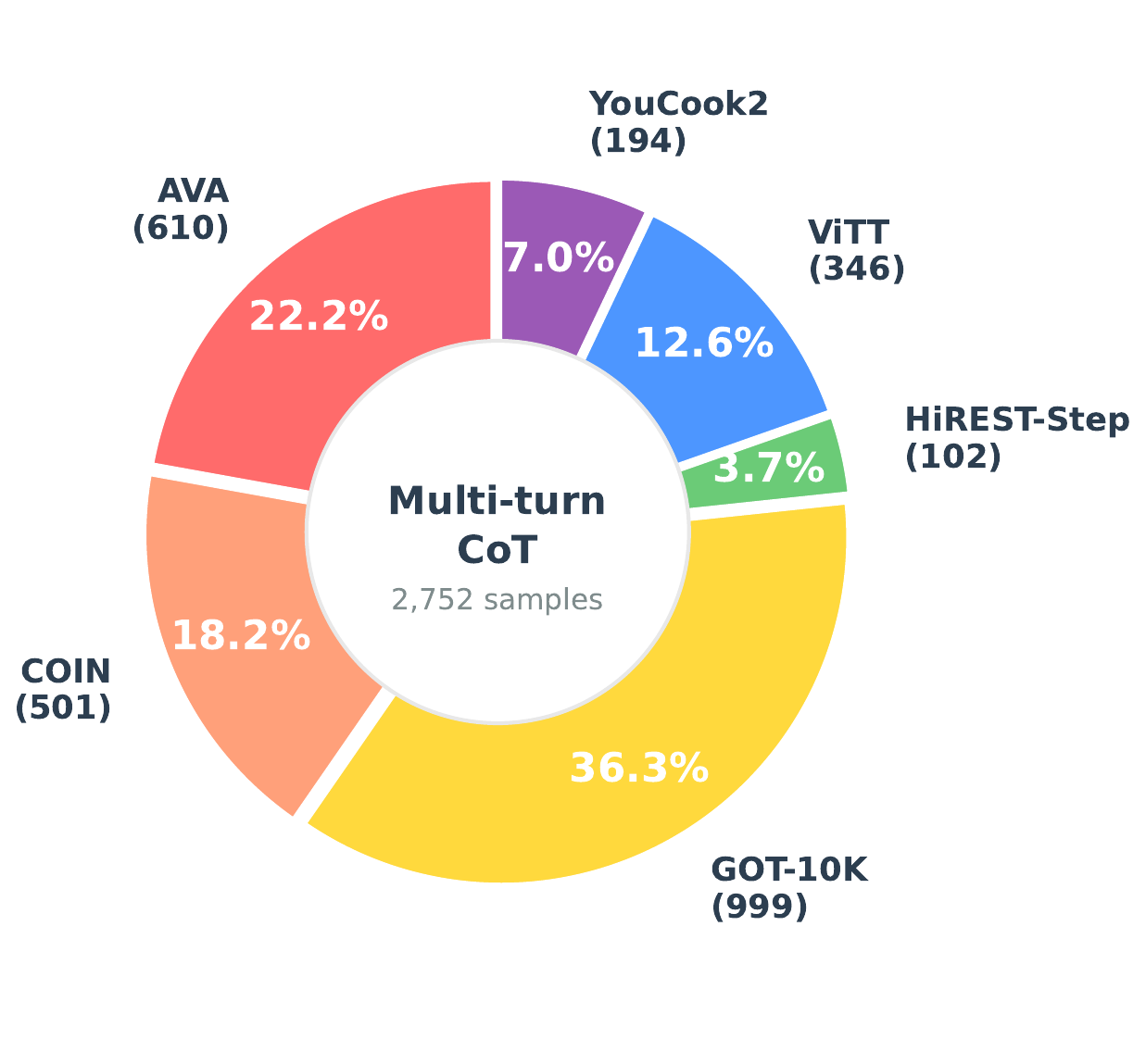}
    \par\small (b) Multi-round CoT.
  \end{minipage}\hfill
  \begin{minipage}[t]{0.32\textwidth}
    \centering
    \includegraphics[width=\linewidth]{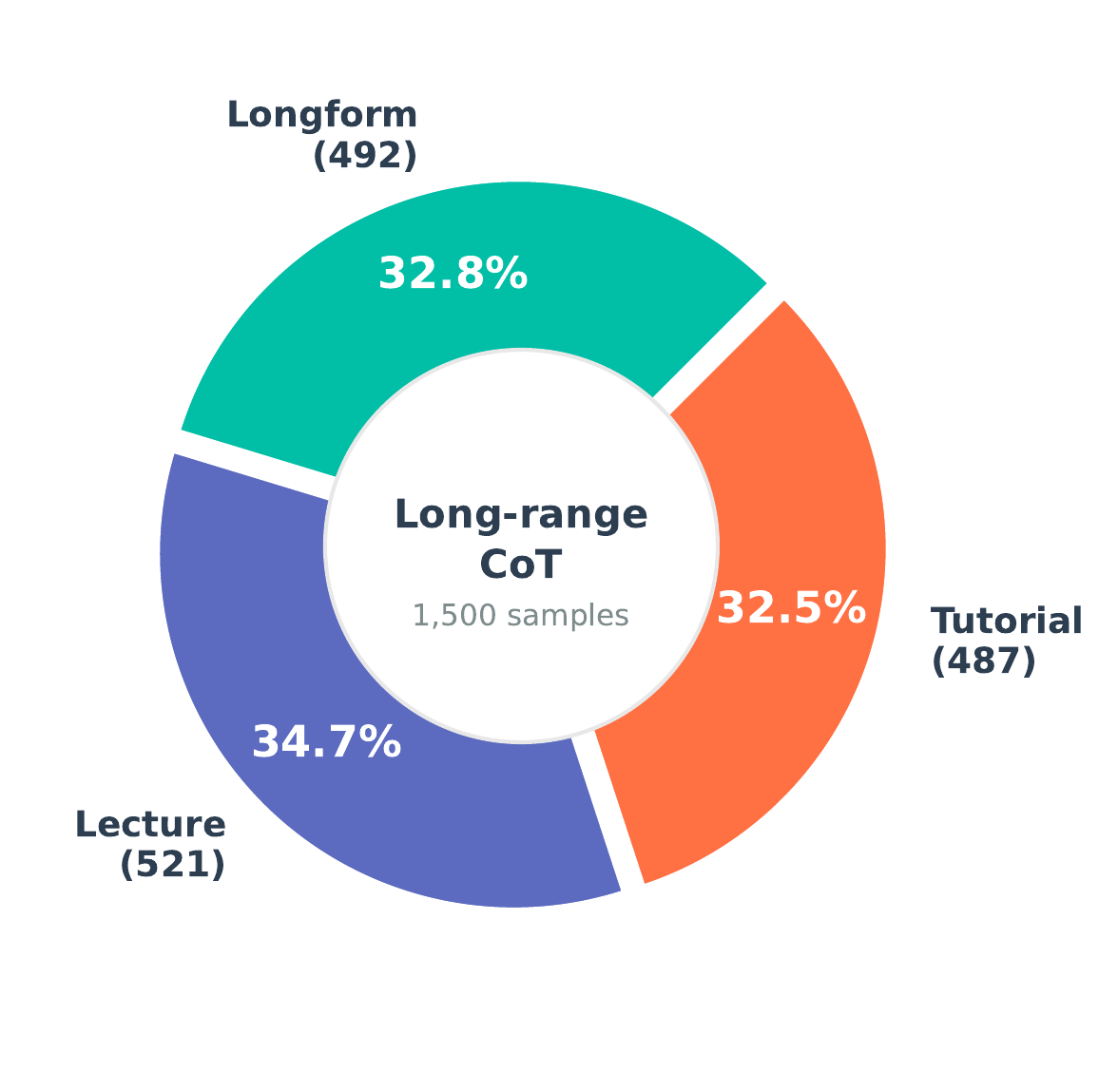}
    \par\small (c) Long-range CoT.
  \end{minipage}
  \caption{\textbf{Dataset composition}}
  \label{fig:data-distributions}
\end{figure*}

\subsection{Overall Dataset Composition}
\label{app:data-composition}

Figure~\ref{fig:data-distributions} visualizes the three-stage data composition and its alignment with our training objectives: Stage~1 for single-round memory writing and answering, Stage~2 for multi-round consistency, and Stage~3 for long-horizon recall, uncertainty handling, and distractor robustness.

\noindent\textbf{VideoChatOnline-IT Source Pool.}
Stage~1 and Stage~2 are both constructed from VideoChatOnline-IT, a video instruction-tuning corpus for online video understanding. It unifies multiple temporally grounded video QA sources in a streaming-style format, making it a suitable source pool for our pseudo streaming chain-of-thought (CoT) construction. We use different subsets for different training objectives. Stage~1 targets single-round streaming CoT and uses HiREST, DiDeMo, and QuerYD. Stage~2 targets multi-round streaming CoT and uses AVA, COIN, GOT-10K, HiREST-Step, ViTT, and YouCook2. Detailed statistics are reported in Table~\ref{tab:single-cot} and Table~\ref{tab:multi-turn-detail}.

\noindent\textbf{Stage 1: Single-round CoT Statistics.}
\label{app:single-cot}
Stage~1 focuses on training the model to write segment-level memory notes and answer a single streaming question grounded in the observed prefix. Table~\ref{tab:single-cot} reports the breakdown by source subset.

\begin{table}[!htbp]
  \caption{\textbf{Stage 1 detailed statistics.}}
  \label{tab:single-cot}
  \centering
  \small
  \begin{tabular}{lrr}
    \toprule
    \textbf{Subset} & \textbf{Samples} & \textbf{Proportion (\%)} \\
    \midrule
    HiREST & 352 & 6.8 \\
    DiDeMo & 4023 & 78.0 \\
    QuerYD & 785 & 15.2 \\
    \midrule
    Total & 5,160 & 100.0 \\
    \bottomrule
  \end{tabular}
\end{table}

\noindent\textbf{Stage 2: Multi-round CoT Statistics.}
\label{app:multi-turn-cot}
Stage~2 trains multi-turn consistency: later answers must reuse earlier segment-level memory notes without peeking into future segments. Table~\ref{tab:multi-turn-detail} reports detailed statistics, including the number of segments and questions aggregated from the underlying sources.

\begin{table}[!htbp]
  \caption{\textbf{Stage 2 detailed statistics.}
  Segments counts the number of segment units after stream segmentation, and Questions counts the number of question turns in the synthesized dialogues. Avg. Segs. and Avg. Qs. denote the average numbers of segments and questions per sample.}
  \label{tab:multi-turn-detail}
  \centering
  \small
  \begin{tabular}{lrrrrr}
    \toprule
    \textbf{Subset} & \textbf{Samples} & \textbf{Segments} & \textbf{Questions} & \textbf{Avg. Segs.} & \textbf{Avg. Qs} \\
    \midrule
    AVA & 610 & 6,136 & 1,902 & 10.06 & 3.12 \\
    COIN & 501 & 2,793 & 1,109 & 5.57 & 2.21 \\
    GOT-10K & 999 & 2,830 & 2,820 & 2.83 & 2.82 \\
    HiREST-Step & 102 & 320 & 422 & 3.14 & 4.14 \\
    ViTT & 346 & 1,090 & 1,436 & 3.15 & 4.15 \\
    YouCook2 & 194 & 630 & 824 & 3.25 & 4.25 \\
    \midrule
    Total & 2,752 & 13,799 & 8,513 & 5.01 & 3.09 \\
    \bottomrule
  \end{tabular}
\end{table}

\noindent\textbf{Stage 3: Long-range CoT Statistics and Retrieval Keywords.}
\label{app:stage3}
Stage~3 targets long-horizon streaming behaviors on long videos collected from YouTube. We use 500+ keywords to retrieve candidate long videos, covering procedural workflows (tutorial), explanatory content (lecture), and continuous recordings (longform). Table~\ref{tab:stage3-detail} reports the resulting dataset statistics by category, and Table~\ref{tab:yt-queries} lists representative query examples used for video retrieval.

\begin{table}[t]
  \caption{\textbf{Stage 3 statistics by category.}}
  \label{tab:stage3-detail}
  \centering
  \small
  \begin{tabular}{lrrrrr}
    \toprule
    \textbf{Category} & \textbf{Samples} & \textbf{Segments} & \textbf{Questions} & \textbf{Avg. Segs.} & \textbf{Avg. Qs} \\
    \midrule
    Tutorial & 487 & 5,631 & 2,017 & 11.56 & 4.14 \\
    Lecture  & 521 & 4,532 & 1,893 & 8.70  & 3.63 \\
    Longform & 492 & 5,536 & 2,090 & 11.25 & 4.25 \\
    \midrule
    Total & 1,500 & 15,699 & 6,000 & 10.47 & 4.00 \\
    \bottomrule
  \end{tabular}
\end{table}

\begin{table}[H]
\caption{\textbf{Core constraints for pseudo streaming CoT generation.}}
\label{tab:prompt-constraints}
\centering
\small
\begin{tabular}{p{0.08\textwidth}p{0.84\textwidth}}
\toprule
\textbf{ID} & \textbf{Constraint} \\
\midrule
A & \textbf{Strict one-to-one alignment:} exactly one output chunk per input unit, preserving order. \\
B & \textbf{No empty segment:} each segment must update/maintain grounded state based on visual evidence. \\
C & \textbf{No future information:} segment reasoning uses only current and past units; QA uses evidence up to its timestamp. \\
D & \textbf{No answer leakage:} do not reveal reference answers in reasoning; copy the answer only in the final Answer field. \\
E & \textbf{Video-grounded only:} rely on provided frames/timestamps/metadata; avoid external assumptions. \\
F & \textbf{Streaming quality:} emphasize boundary cues and conservative state updates across segments. \\
G & \textbf{Question awareness:} track unanswered questions and prioritize collecting relevant evidence online. \\
\bottomrule
\end{tabular}
\end{table}

\begin{table}[t]
  \caption{\textbf{Representative YouTube search queries used for Stage~3 video retrieval.}}
  \label{tab:yt-queries}
  \centering
  \small
  \setlength{\tabcolsep}{4pt}
  \renewcommand{\arraystretch}{1.05}
  \begin{tabular}{p{0.44\textwidth}p{0.42\textwidth}}
    \toprule
    \textbf{Query keyword} & \textbf{Video type} \\
    \midrule
    \multicolumn{2}{l}{\textbf{Tutorial}} \\
    \texttt{sourdough bread tutorial} & Complete bread-making workflow \\
    \texttt{furniture restoration} & Furniture restoration project \\
    \texttt{oil painting tutorial} & Oil painting step-by-step tutorial \\
    \texttt{car repair tutorial complete} & Full car repair walkthrough \\
    \texttt{sewing tutorial complete} & Complete sewing tutorial \\
    \texttt{woodworking project tutorial} & Woodworking project tutorial \\
    \texttt{pottery making tutorial} & Pottery making process \\
    \texttt{knife making tutorial} & Knife forging tutorial \\
    \addlinespace[0.3em]

    \multicolumn{2}{l}{\textbf{Lecture}} \\
    \texttt{machine learning lecture} & Machine learning course lecture \\
    \texttt{organic chemistry lecture} & Organic chemistry lecture \\
    \texttt{quantum mechanics lecture} & Quantum mechanics lecture \\
    \texttt{algorithm course full} & Full algorithm course \\
    \texttt{system design lecture} & System design lecture \\
    \texttt{neuroscience lecture} & Neuroscience lecture \\
    \texttt{deep learning tutorial} & Deep learning lecture/tutorial \\
    \texttt{computer vision lecture} & Computer vision lecture \\
    \addlinespace[0.3em]

    \multicolumn{2}{l}{\textbf{Longform}} \\
    \texttt{hiking trail complete} & Full hiking trail recording \\
    \texttt{train journey scenic} & Scenic train journey \\
    \texttt{safari wildlife documentary} & Safari wildlife documentary \\
    \texttt{Tokyo walking tour} & City walking tour (Tokyo) \\
    \texttt{northern lights footage} & Northern lights raw footage \\
    \texttt{coral reef documentary} & Coral reef documentary \\
    \texttt{mountain climbing documentary} & Mountain climbing documentary \\
    \texttt{storm chasing footage} & Storm chasing raw footage \\
    \bottomrule
  \end{tabular}
\end{table}

\subsection{Pseudo Streaming CoT Generation Principles}
\label{app:synthesis-principles}

We synthesize pseudo streaming CoT annotations to match the streaming protocol in Sec.~\ref{sec:paradigm}. A key constraint is strict alignment: for a stream with $S$ segments and $Q$ questions, the synthesized output must contain exactly $S+Q$ generated items, one for each interleaved unit and in temporal order. Table~\ref{tab:prompt-constraints} summarizes the core constraints enforced during CoT synthesis.

\noindent\textbf{Prompt template.}
For reproducibility, we provide the complete prompt used to synthesize pseudo streaming CoT. We use special tokens to delimit streaming units: \texttt{$<$EOS$>$} ends an input segment unit, \texttt{$<$EOQ$>$} ends an input question unit (and also the corresponding QA output), and \texttt{$<$EOT$>$} ends a generated segment-reasoning chunk. In training data, we keep only essential delimiters to reduce overfitting to superficial formatting.

\begin{PromptBox}
general_prompt = '''
You are a pseudo streaming Video Chain-of-Thought (CoT) generator.
You will receive the FULL input at once (all video segments + all questions + reference answers),
but you MUST generate reasoning that looks as if you processed the video incrementally, in time order.
====================
1) CORE CONSTRAINTS
====================
(A) STRICT ONE-TO-ONE ALIGNMENT (mandatory)
- The input is a chronological sequence of interleaved units.
- Video segment units end with <EOS>.
- Question units end with <EOQ>.
- You MUST output exactly ONE reasoning chunk per input unit.
- Output order must exactly match input order.
- Total output chunks = (#segment units + #question units).
- Do NOT merge units.
- Do NOT split units.
- Do NOT reorder units.
(B) EVERY SEGMENT REQUIRES REASONING (no empty segments)
- For every video segment unit, produce meaningful reasoning grounded in that segment.
- If no task-relevant change occurs, explicitly state that the scene, action, or tracked state remains stable.
- Avoid meta statements; focus on video evidence and state continuity.
(C) pseudo streaming / NO FUTURE INFORMATION
- Although you see the full input, behave as if you only know information up to the current unit.
- In [SEG k THINK], you may ONLY use evidence from segment k and all earlier segments.
- NEVER use or hint at information that appears only in future segments, questions, or answers.
- In [Q j THINK], reason ONLY with evidence available up to the question timestamp t.
(D) NO ANSWER LEAKAGE
- Each question includes a Reference Answer (for alignment only).
- NEVER reveal, paraphrase, or hint at any reference answer in any segment reasoning.
- In [Q j THINK], output the final answer ONLY inside the Answer field.
- The Answer MUST be copied EXACTLY and VERBATIM from the Reference Answer.
- Do NOT leak the answer in the Reasoning field.
(E) VIDEO-GROUNDED ONLY
- Use only visual evidence provided by frames, timestamps, bounding boxes, and metadata.
- Do NOT rely on external knowledge, assumptions, or commonsense completion.
(F) STREAMING VIDEO REASONING QUALITY
Each segment reasoning should:
- Describe what is visually observed or confirmed in this segment.
- Emphasize continuity, change, or boundary cues (start / end / ongoing).
- Update internal task-specific state clearly and conservatively.
(G) QUESTION AWARENESS (when applicable)
- If one or more questions have appeared:
  - Maintain an "Active Question": the earliest question that has not been answered.
  - Segment reasoning should prioritize collecting evidence relevant to the Active Question.
- If NO question has appeared yet:
  - Focus ONLY on understanding the video stream itself: scene setup, object/person continuity,
    motion patterns, emerging actions.
  - Do NOT speculate about future questions.
====================
2) INPUT FORMAT
====================
Units are interleaved in chronological order.
Video segment unit:
[SEG k | time = start-end | frames = ... | optional: bboxes / ids / actions] <EOS>
Question unit:
[Q j | t = timestamp]
Question: ...
Reference Answer: ...
<EOQ>
Notes:
- Timestamps may not start at 0; follow the provided time system exactly.
- Reference Answers are provided for alignment only.
====================
3) OUTPUT FORMAT (STRICT)
====================
For each video segment unit:
[SEG k THINK]
Focus: (either the Active Question in <= 15 words, OR "video understanding (no question yet)")
Evidence from this segment: (2-5 sentences, strictly video-grounded)
State update: (1-3 sentences, task-specific state or continuity)
<EOT>

For each question unit:
[Q j THINK]
Reasoning: (2-6 sentences, justify the answer using ONLY evidence available up to time t;
           you may reference segment indices/timestamps, but MUST NOT use future units
           and MUST NOT reveal/paraphrase the Reference Answer in Reasoning)
Answer: (copy the Reference Answer EXACTLY and VERBATIM)
<EOQ>
'''
\end{PromptBox}

\section{Case Studies}
\label{app:Study_Case}

\begin{figure*}[p]
  \centering
  \includegraphics[width=\textwidth,height=0.85\textheight,keepaspectratio]{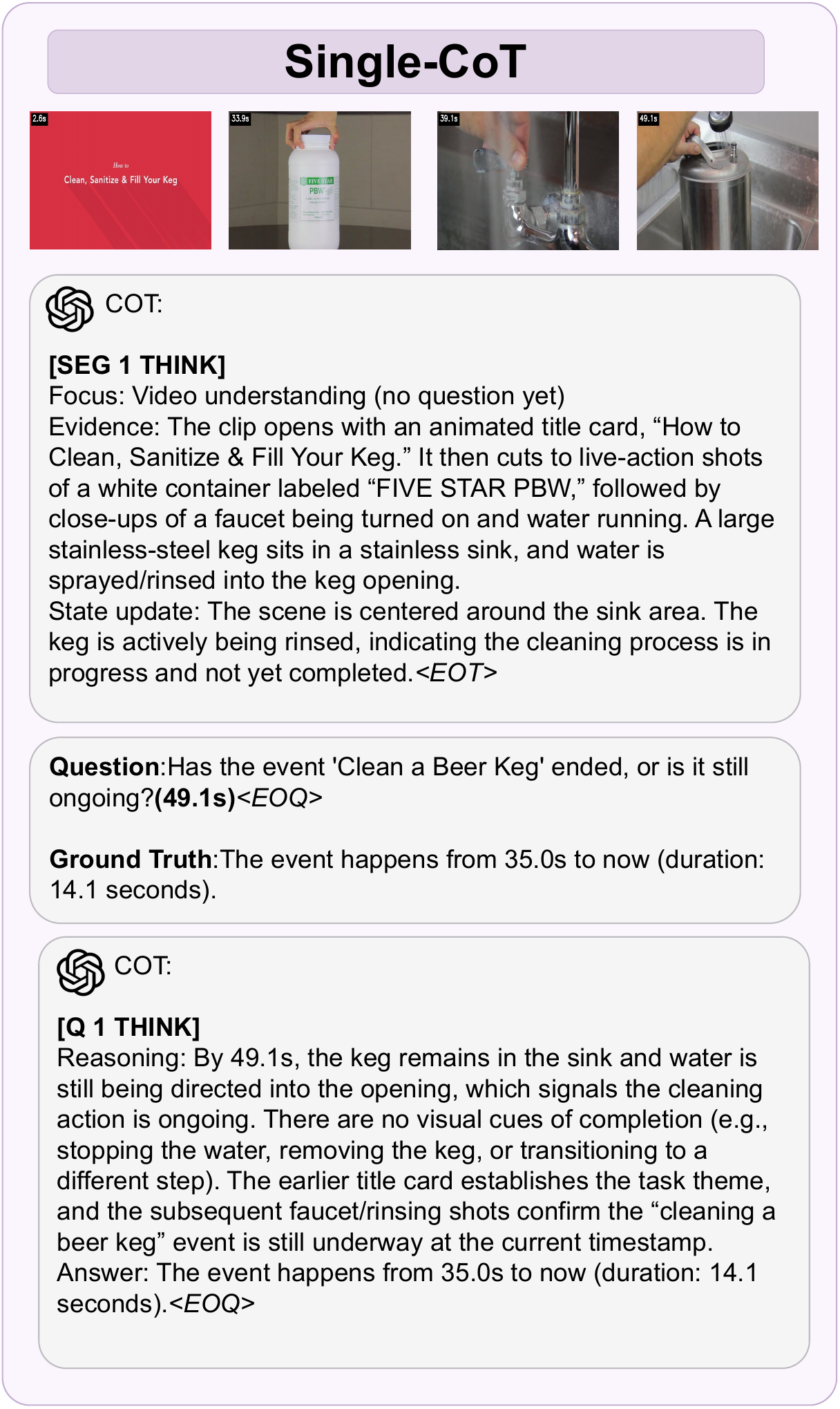}
  \caption{\textbf{Dataset-derived pseudo streaming CoT example under the single-round protocol.}
  A question is asked once after the observed video prefix. The model first writes a segment-level memory note from the incoming frames, identifying the tutorial title card, the cleaning material, the faucet, and the rinsing action around the keg. When queried at 49.1s, it answers using only the accumulated evidence so far and correctly concludes that the event \textbf{Clean a Beer Keg} is still ongoing.}
  \label{fig:qual-single}
\end{figure*}

\begin{figure*}[p]
  \centering
  \includegraphics[width=\textwidth,height=0.85\textheight,keepaspectratio]{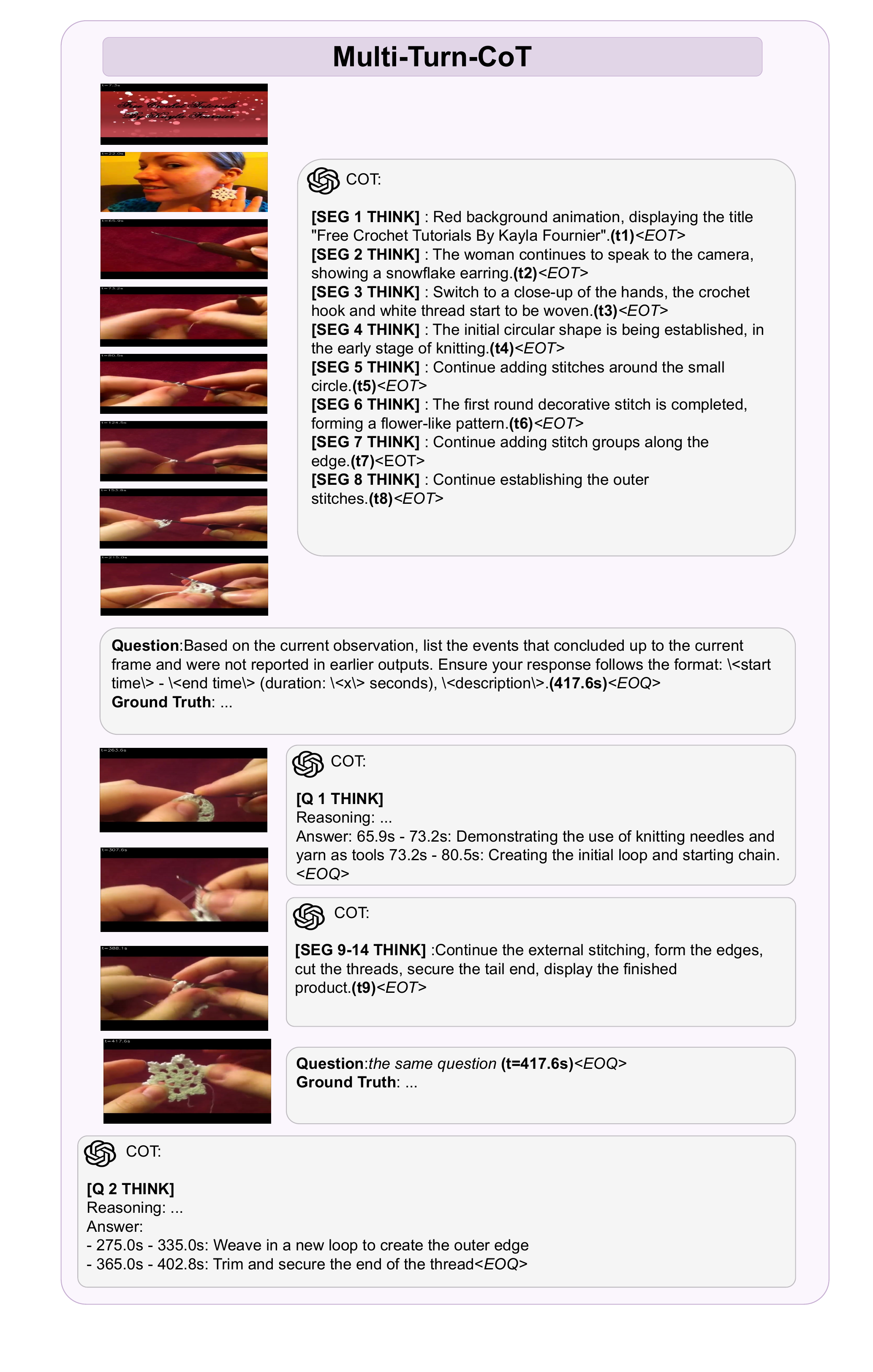}
  \caption{\textbf{Dataset-derived pseudo streaming CoT example under the multi-round protocol.}
 The stream follows a crochet tutorial. Memory notes from Segments 1 to 8 trace the progression from the introduction to close-up hand actions and the gradual formation of the crochet pattern. After the first event-listing question, the model reports only the completed events observed so far. As additional segments arrive, it updates the memory with later stitching and finishing steps. When the same question is asked again at 417.6s, the answer includes only the newly completed events that were not reported earlier, which illustrates incremental reasoning across multiple turns over a continuous stream.}
  \label{fig:qual-multi}
\end{figure*}

\begin{figure*}[p]
  \centering
  \includegraphics[width=\textwidth,height=0.80\textheight,keepaspectratio]{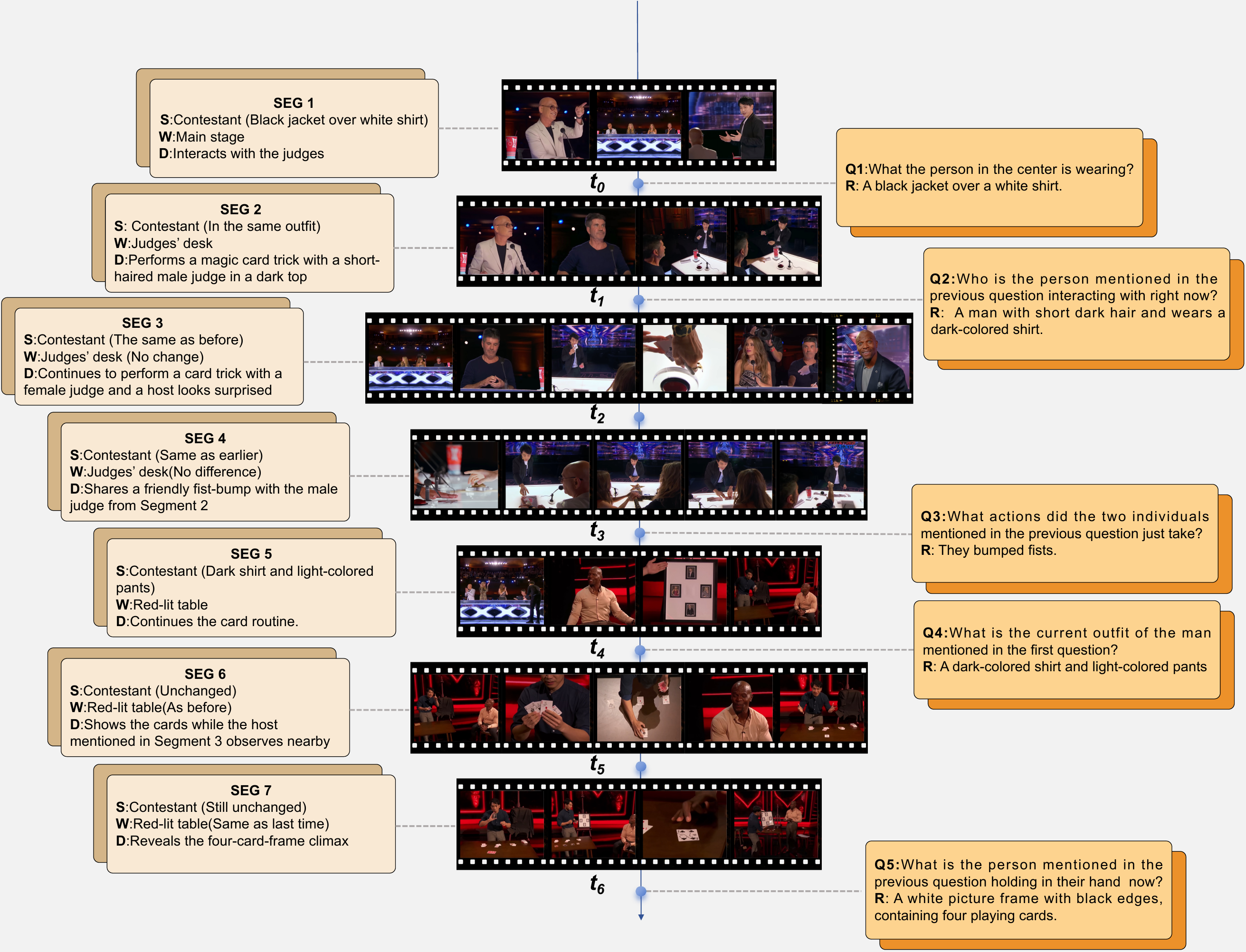}
  \caption{\textbf{Real multi-round streaming example.}
  Left: Segment-level memory notes summarize the contestant's appearance, location, and actions across seven streamed segments. Middle: Representative frames show the progression of a card routine from the main stage to the judges' desk and finally to a red-lit table. Right: Five temporally ordered questions are interleaved with the stream. Questions Q2 to Q5 depend on earlier turns, requiring the model to resolve the previously mentioned person, identify the current interaction partner, recall the recent fist bump, update the contestant's outfit after the scene transition, and track the object held in hand at the end. This example illustrates how persistent segment-level memory supports consistent reasoning across multiple turns over a changing stream.}
  \label{fig:qual-case-study}
\end{figure*}

This section presents three qualitative examples that complement the quantitative results. \Cref{fig:qual-single} shows a dataset-derived pseudo streaming CoT example under the single-round protocol. \Cref{fig:qual-multi} shows a dataset-derived pseudo streaming CoT example under the multi-round protocol. \Cref{fig:qual-case-study} shows a real multi-round streaming example and illustrates how segment-level memory supports cross-turn reference resolution and temporal state tracking.

\section{Error Analysis}
\label{app:error}

\begin{figure*}[p]
  \centering
  \includegraphics[width=\textwidth,height=0.80\textheight,keepaspectratio]{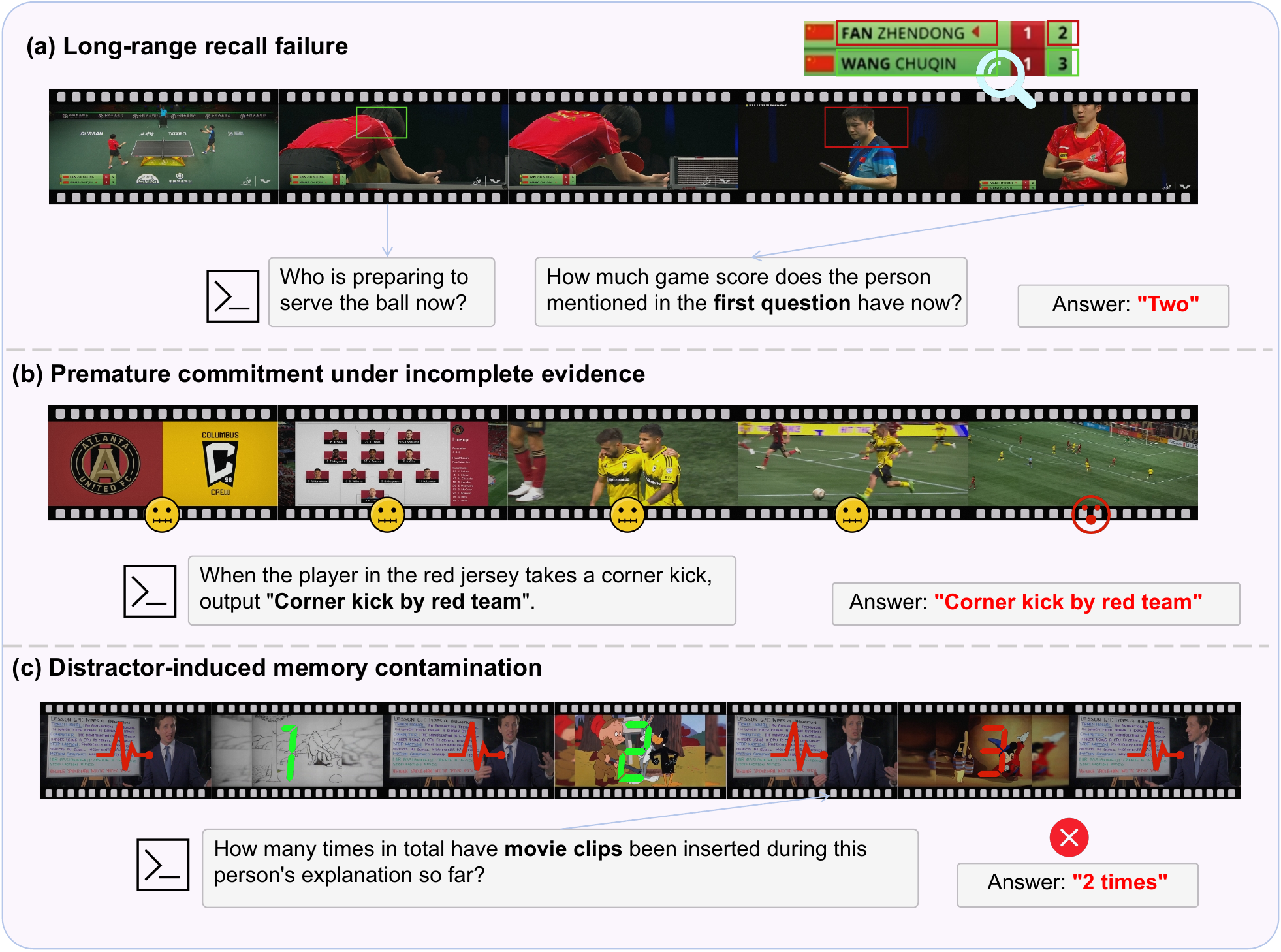}
  \caption{\textbf{Residual failure cases in multi-turn streaming reasoning.}
  \textbf{(a) Long-range recall failure.} The model fails to recover an early fine-grained identity cue across turns. As illustrated in the top row, the \textcolor{green}{green box} marks the correct person referred to in the first question, whereas the \textcolor{red}{red box} marks the person incorrectly retrieved by the model in the later question, indicating that the coarse event context is retained but the specific identity association is lost.
  \textbf{(b) Premature commitment under incomplete evidence.} The middle row shows that, at the queried moment, the decisive visual evidence for whether the red-jersey player is taking a corner kick has not yet become sufficiently clear. Instead of deferring its response until the event is visually confirmed, the model commits to a specific answer too early.
  \textbf{(c) Distractor-induced memory contamination.} The bottom row shows repeated alternation between the speaker and inserted movie clips. Because these inserted clips appear interleaved with the main explanation, the model is distracted by the intervening visual switches and produces an incorrect insertion count.
  These failures are residual edge cases of the three long-horizon behaviors explicitly targeted in Stage~3 training, namely long-term evidence retention, uncertainty handling, and distractor robustness.}
  \label{fig:error_cases}
\end{figure*}

Although Stage~3 substantially improves long-horizon streaming reasoning, representative residual failures still remain in challenging multi-turn settings. Consistent with the three long-horizon behaviors explicitly targeted in Stage~3 training---long-term evidence retention, uncertainty handling, and distractor learning, as described in Sec.~\ref{sec:method:data}---we observe three recurring error patterns. First, the model may retain the coarse event trace while forgetting an early fine-grained attribute, such as which object, side, or entity was involved. This is consistent with our segment-level memory design: compact memory notes support long-range access but can still over-compress details over long temporal gaps. Second, under incomplete evidence, the model may commit to a specific hypothesis too early rather than deferring judgment until decisive visual evidence appears, reflecting a residual limitation of the uncertainty-handling objective in Sec.~\ref{sec:method:data}. Third, later retrieval can still be corrupted by visually salient but task-irrelevant segments, causing recent distractors to override the true earlier evidence. This aligns with the ablation results in Table~\ref{tab:memory_ablation}, which show that memory notes help stabilize retrieval but remain limited by the quality of written evidence and incoming visual context. As shown in Fig.~\ref{fig:error_cases}, these failures are residual edge cases of (a) long-range recall failure, (b) premature commitment under incomplete evidence, and (c) distractor-induced memory contamination in streaming multi-turn reasoning.

\clearpage

\clearpage

\clearpage

\end{document}